\documentclass[10pt,twocolumn,letterpaper]{article}

\usepackage{iccv}
\usepackage{times}
\usepackage{epsfig}
\usepackage{graphicx}
\usepackage{amsmath}
\usepackage{amssymb}
\usepackage{todonotes} 
\usepackage{comment}
\usepackage{pdfpages}

\usepackage{algorithm}
\usepackage{algorithmic}
\usepackage{tabularx}
\usepackage{flafter}
\usepackage{siunitx}
\usepackage{todonotes} 
\usepackage{booktabs}

\usepackage[export]{adjustbox}
\usepackage{hyperref}
\usepackage[accsupp]{axessibility} 

\iccvfinalcopy 

\def\iccvPaperID{9284} 

\ificcvfinal\fi

\begin{document}

\title{Deformable Neural Radiance Fields using RGB and Event Cameras}

\author{ Qi Ma\textsuperscript{1}\space\space\space\space Danda Pani Paudel\textsuperscript{1,3}\space\space\space\space Ajad Chhatkuli\textsuperscript{1}\space\space\space\space Luc Van Gool\textsuperscript{1,2,3}\\
\textsuperscript{1}Computer Vision Lab, ETH Zurich\space\space\space\space \textsuperscript{2}VISICS, ESAT/PSI, KU Leuven\space\space\space\space \textsuperscript{3}INSAIT, Sofia University  }


\maketitle
\ificcvfinal\thispagestyle{empty}\fi

\begin{abstract}
   Modeling Neural Radiance Fields for fast-moving deformable objects from visual data alone is a challenging problem. A major issue arises due to the high deformation and low acquisition rates. To address this problem, we propose to use event cameras that offer very fast
   acquisition of visual change in an asynchronous manner. In this work, we develop a novel method to model the deformable neural radiance fields using RGB and event cameras. The proposed method uses the asynchronous stream of events and calibrated sparse RGB frames. In our setup, the camera pose at the individual events --required to integrate them into the radiance fields-- remains unknown. Our method jointly optimizes these poses and the radiance field. This happens efficiently by leveraging the collection of events at once and actively sampling the events during learning. Experiments conducted on both realistically rendered graphics and real-world datasets demonstrate a significant benefit of the proposed method over the state-of-the-art and the compared baseline.     
   This shows a promising direction for modeling deformable neural radiance fields in real-world dynamic scenes. We release our code at: \href{https://qimaqi.github.io/DE-NeRF.github.io/}{https://qimaqi.github.io/DE-NeRF.github.io/}

\end{abstract}

\begin{figure}
\begin{center}
\includegraphics[width=1.0\linewidth]{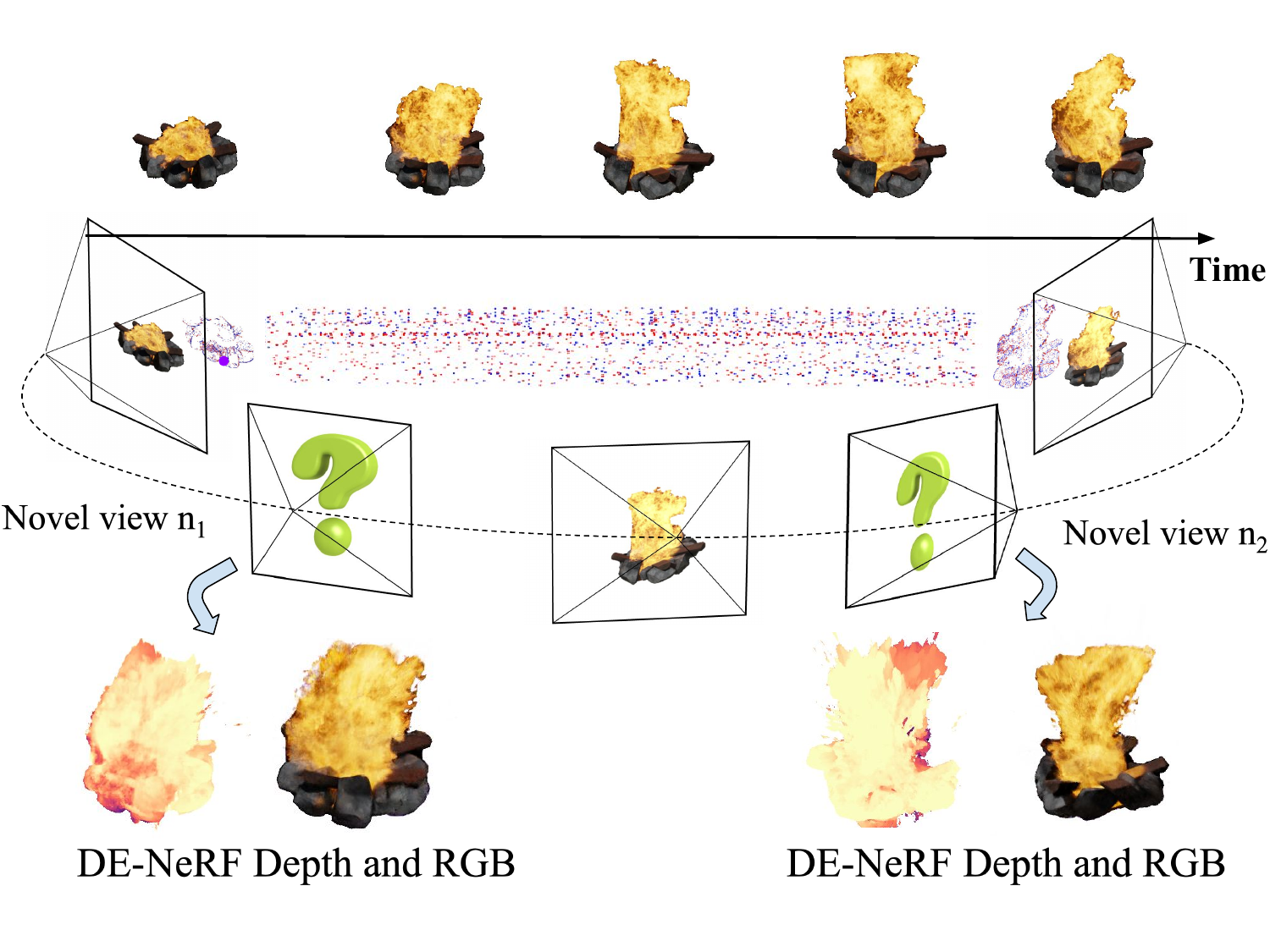}
\end{center}
  \caption{Our framework takes the aligned frames and events captured by a dual RGB-Event camera setup as input. Our method captures fast-moving objects and is capable of rendering a free-viewpoint representation at given timestamps. The figures show the flames' reconstruction with high quality and correct geometry. }
\label{fig:problem}
\end{figure}

\section{Introduction}
Neural Radiance Fields (NeRFs) have shown great success in synthesizing photorealistic images by implicitly representing rigid 3D scenes. Modeling non-rigid
scenes in such manner is a much more difficult task. Recently, several methods have been proposed to model dynamic neural radiance fields. They aim to model
rather slowly deforming radiance fields ~\cite{gao2021dynamic, park2021nerfies, park2021hypernerf, gao2022monocular, weng2022humannerf}. The slow deformation assumption is insufficient in scenarios
involving fast-moving objects or, equivalently, low frame-
rate cameras. In other words, the existing methods cannot capture fast-deforming radiance fields due to the limited
frame rate of RGB cameras. To address this problem, we
propose to add an event camera that provides
information about the radiance change asynchronously.

Event cameras capture radiance changes, also in the presence of fast motions. However, harnessing this benefit comes with its own challenges, mainly due to (i) the unknown absolute radiance at the event location and (ii) the unknown pose of the camera at the time of the event. The former challenge can be addressed by using a hybrid system of RGB and event cameras. We address the latter challenge of pose determination with a novel method.

Previous methods to deal with the pose of event cameras either do not treat the events to be asynchronous~\cite{kim2016real,kim2014simultaneous} or assume that the event camera's pose is known at all times~\cite{rudnev2022eventnerf}. We advocate that the event streams must be treated asynchronously to maximally utilize their temporal precision, in keeping with earlier work~\cite{schaefer2022aegnn}. On the other hand, we argue that the assumption of known poses for all asynchronous events is simply impractical. Instead, we assume that only the poses of subsequent RGB frames are known. The poses of the events are derived from their associated time stamps, by learning to map time to the evolving camera poses. During this process, the known poses of the RGB frames and the non-rigid deformation prior of the scene under investigation are jointly utilized.

In this work, we use moving calibrated stereo of RGB and event cameras. Using the known poses of sparse RGB frames only, we want to model the 3D radiance field of deformable objects. To the best of our knowledge, there are thus far no methods leveraging event cameras to model deformable neural radiance fields. Therefore, we first establish a baseline method -- which we refer to as DE-baseline -- inspired by two notable works on deformable NeRF~\cite{park2021nerfies} and event-based NeRF~\cite{rudnev2022eventnerf}. Later, we propose a novel method that significantly improves this baseline. The proposed method learns to map the time stamp of an event to a camera pose such that each event's ray can be backprojected to the 3D space, without requiring the continuous pose of the asynchronous events. The main idea of this paper is then to constrain the radiance field using the measured events. 
To do so, we re-create the events solely from the radiance field. Any error due to mismatches between re-created and measured events is backpropagated to supervise the implicit radiance field represention. This radiance field is augmented by sparse and calibrated RGB image frames. The major contributions of this paper are as follows,
\begin{itemize}
   \item We show the benefit of using event cameras to model the deformable neural radiance fields for the first time. 

   \item We develop a novel method that learns the continuous pose of event cameras which is robust also to inaccurate RGB poses, exploits a collection of events at once, and performs active sampling to maximally utilize the asynchronous event streams. 
   
   \item The proposed method significantly outperforms existing methods and our baseline on both realistically rendered, but artificial scenes and on real-world datasets.
\end{itemize}

\section{Related Works}
\paragraph{Dynamic NeRF:}
Dynamic NeRFs~\cite{pumarola2021d,park2021nerfies,park2021hypernerf,song2023nerfplayer} address the challenging problem of representing static, dynamic or non-rigid scenes using radiance fields~\cite{mildenhall2021nerf}. Several works on dynamic NeRF use model-based approaches, e.g., representing human body, hands or faces~\cite{weng2022humannerf,xiu2022icon,Ye2023}. Model-free approaches on the other hand, learn a generic deformation function in order to represent non-rigid camera projections or 3D scenes, which is also our interest in this work. Early work D-NERF~\cite{pumarola2021d} uses a chosen canonical view to map deformed scenes using a time conditioned function represented by Multi-layer Perceptrons (MLPs). Non-rigidNERF~\cite{tretschk2021non} instead deforms the viewing rays and thus the projections instead of the 3D surface, thus, the approach does not directly provide the 3D of the deformed scene. Nerfies~\cite{park2021nerfies} train a time conditioned deformation function much like D-NERF~\cite{pumarola2021d}, albeit with an unknown canonical template-based neural field representation~\cite{Zheng_2021_CVPR}. Furthermore, it also regularizes the deformation field using a coarse-to-fine strategy. As the deformation is defined on 3D space, it can effectively render depths of the non-rigid scene at different time values. 
HyperNERF~\cite{park2021hypernerf} introduces shape embeddings in higher dimensions in order to handle topological changes. A very recent work~\cite{song2023nerfplayer} trains NERF for streambale rendering while representing static, rigid and non-rigid scene elements separately.

\paragraph{Event cameras for 3D Vision.}
Event cameras for 3D reconstruction and camera tracking were presented in \cite{kim2016real}, based on a probabilistic framework for disparity estimation. Later, \cite{bryner2019event} addressed camera tracking through a generative modeling of events and maximum likelihood estimation of camera motion. Other contributions have proposed solutions for direct sparse~\cite{hidalgo2022event} and stereo~\cite{zhou2021event} visual odometry. \cite{rebecq2018emvs} solves semi-dense multi-view stereo from known poses -- by exploiting object silhouettes seen by a  moving camera. Similarly, \cite{zhou2018semi} tackles semi-dense stereo-based 3D reconstruction by also solving for the camera motion. \cite{baudron2020e3d} reconstructs shapes as a shape from silhouette problem and handles single object reconstruction through synthetic data training. \cite{wang2022evac3d} presents a shape from silhouette solution with high quality. Recently, ~\cite{xue2022event} solves non-rigid 3D reconstruction from contours using event cameras. These have also been used for 3D hand pose analysis \cite{rudnev2021eventhands}.

\paragraph{Event NeRF.}
Unlike traditional approaches for 3D reconstruction, NERF-based 3D reconstruction in event cameras is largely under-explored. The generative model-based view synthesis together with surface density estimation in NERF requires highly accurate camera poses and careful optimization, thus rendering its application in event cameras highly challenging. Recent work Event-NERF~\cite{rudnev2022eventnerf} makes use of a single colour event camera in order to optimize radiance fields, while assuming that the background colour is known in advance. It introduces random temporal window sampling in order to provide diverse supervision. E-NERF~\cite{klenk2023nerf} presents a NERF method for event frames or events with RGB images in synthetic scenes. The method proposes a normalized loss function in order to handle varying contrast threshold of event cameras. In particular, the method effectively solves deblurring of images using events. Another parallel work Ev-NERF~\cite{hwang2023ev} also proposes an event-based NERF method, which uses a threshold-bound loss in order to address the lack of RGB images. The event-to-frame method such as E2VID~\cite{rebecq2019high} was employed easily with frame-based NeRF, revealing poor performance that aligns with the findings in our work. All of the previous methods consider the scene to be static, with various assumptions on the contrast threshold of the event camera~\cite{hwang2023ev,rudnev2022eventnerf}. A natural question is thus, can event cameras be used to construct NERF to obtain high quality 3D reconstruction with dynamic objects or scenes, where events can provide a significant edge over conventional cameras? If so, how can we tackle highly challenging non-rigid scenes not addressed by any previous methods? In the following sections we answer these two questions with our proposed method and experiments.

\section{Events in the Radiance Field}
We represent the pose of the events as a function of time $\mathsf{P}(t)$. At any time $t$, the 6DoF pose is parameterized by the screw axis $\mathsf{S}=(r(t);v(t)) \in \mathbb{R}^{6} $ where the rotation matrix and translation vectors can be recovered by Rodrigues's formula~\cite{lynch2017modern}. Without loss of generality, we avoid representing the pose of the RGB camera separately. Whenever needed, the RGB camera's pose is related to $\mathsf{P}(t)$ using the known camera extrinsic parameters between the RGB and event cameras. A tuple $\mathbf{e} = (x,t)$ is an event triggered at 2D location $x$ and time $t$. An event camera measures a set of such tuples, say $\mathcal{E}= \{\mathbf{e}_i\}$.  At a sparse set of time stamps, say $\mathcal{T}_s=\{t_j\}$, RGB images with known pose $\mathcal{R}=\{(\mathcal{I}_j, \mathsf{P}(t_j))\}$ are recorded.
We are now interested to model the deformable radiance field only using $\mathcal{R}$ and $\mathcal{E}$. 

\begin{figure}[h]
\includegraphics[width=0.48\textwidth]{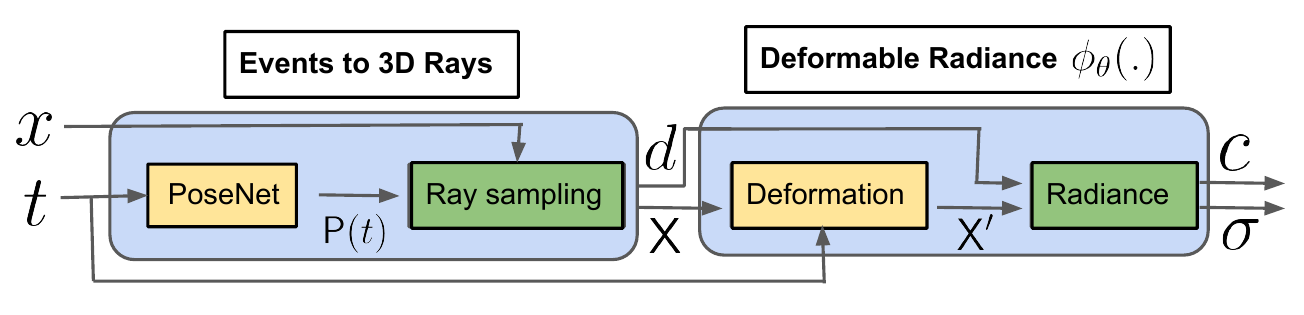}        
   \caption{\textbf{Events to radiance mapping.} The 2D points $x$ and time $t$  are first mapped to the 3D points $\mathsf{X}$ along the viewing direction $d$, using pose $\mathsf{P}(t)$. Each sampled point is mapped to the canonical space by deformation and decoded into color $c$ and density $\sigma$.}
   \label{fig:eventOverview}
\end{figure}

We model the radiance field using the implicit neural representation, with the help of a neural network ${\phi_\theta:(\mathsf{X},\mathsf{d},t)\rightarrow (\mathsf{c},\sigma)}$ parameterized by $\theta$. Here, any 3D point $\mathsf{X}$, in the world coordinate frame, seen from the viewing direction $\mathsf{d}$ at time $t$ is mapped to its color $\mathsf{c}$ and density $\sigma$. The goal of this paper is to learn $\theta$( from $\mathcal{E}$ and $\mathcal{R}$ with the object deformation prior. We embed the deformation prior in the network architecture. In the following, we first present the role of an event $\mathbf{e}$ in learning $\theta$. An overview of mapping events to radiance is illustrated in Figure~\ref{fig:eventOverview}.

\subsection{Mapping Events to 3D Rays}

\begin{figure}[h]
\centering
\includegraphics[width=0.42\textwidth]{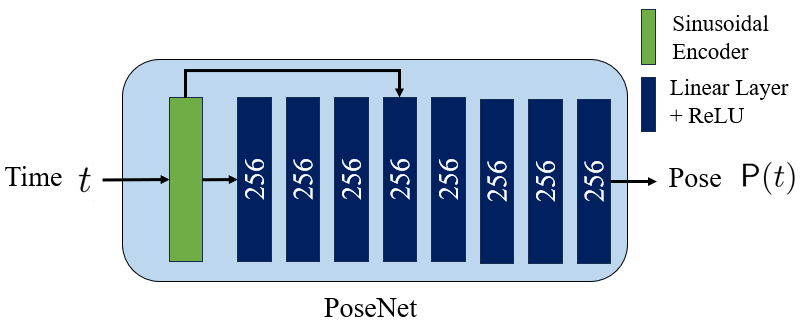}        
   \caption{\textbf{Time to Pose mapping.} We exploit the implicit neural representation to optimize the camera pose as a continuous function of time.  The time $t\in \mathbb{R}$ is firstly normalized to $[-1,1]$ and then pass to sinusoidal encoder with $L=10$ of encoded frequencies per axis. The output of network is mapped to rotation and translation using Rodrigues's formula. }
   \label{fig:posenet}
\end{figure}

\paragraph{Pose from PoseNet.} Every event $\mathbf{e}\in\mathcal{E}$ is first mapped to the corresponding pose $\mathsf{P}(t)$, of the camera at the time when the event was triggered. We realize this mapping using the multi-layer perceptron, \textbf{PoseNet} as shown in Figure~\ref{fig:posenet} which maps time to screw axis representation $(t)\rightarrow(r;v)$. This neural network generates a continuous pose as a function of time, making it very suitable to handle asynchronous events. The knowledge of the camera pose at the event's time and location allows us to backproject the event into the 3D space represented in the world frame. Unlike other Event-based NeRF that employ trajectory interpolation or turntable poses, we address the joint problem of learning neural 3D representation and refining imperfect event poses similar to\cite{lin2021barf} .


\paragraph{Sampling event rays.} 
Once the event is backprojected, a set of points are sampled along the ray, as in the standard setting of NeRF training. Then, a trio of a sampled point, viewing direction and event time is formed to infer its radiance and density. Let the 3D point $\mathsf{X}$, direction $\mathsf{d}$, and time $t$ be such a trio. During inferring radiance and density for this trio, the deformation prior is used in network architecture.

\subsection{Event Rays in the Deformable Radiance Field}
For the deformable radiance field, we assume that there exists a mapping from the deformed surface to a canonical one, as in~\cite{Zheng_2021_CVPR,park2021nerfies}.
Therefore, we first learn to map the 3D point $\mathsf{X}$ observed at time $t$ to its canonical position $\mathsf{X}'$, by learning the inverse deformation field $\omega(\mathsf{X},t)$, such that ${\mathsf{X}'=\mathsf{X}+\omega(\mathsf{X},t)}$. We realize this inverse deformation field using a multi-layer perceptron. The canonical representation $\mathsf{X}'$ is then mapped to the color and density values using another multi-layer perceptron, that additionally receives the viewing direction as input. The arrangement of these two perceptrons, as shown in Figure~\ref{fig:eventOverview} helps us to realize the deformable radiance mapping network ${\phi_\theta:(\mathsf{X},\mathsf{d},t)\rightarrow (\mathsf{c},\sigma)}$, where $\theta$ is the union of parameters of two sub-networks.

\subsection{Rendering Event Ray for Supervision}
Let $\mathcal{I}_e\in\mathcal{R}$ be the nearest available RGB image for any event $\mathbf{e}\in \mathcal{E}$. This nearest association is made by comparing the times in tuple $(x,t)$ and the sparse set of time stamps $\mathcal{T}_s$. We then count the effective number of events $n_e=n_e^p-n_e^n$, for $n_e^p$ positive and $n_e^n$ negative number of events which occurred between the time intervals of  $\mathcal{I}_e$ and  $\mathbf{e}$ acquisitions. The contrast threshold parameter $\tau$ is considered as known. Following the standard volume-rendering~\cite{mildenhall2021nerf} strategy, we render the color $\mathcal{I}_{vr}(\mathbf{e})$ for each event. The rendered color  $\mathcal{I}_{vr}(\mathbf{e})$ is then compared against the  RGB image's color at the event location $\mathcal{I}_e(x)$, while considering the number of intermediate events. More precisely, the event loss for the deformable neural radiance field supervision given by,

\begin{equation}
\mathcal{L}_{event} = \sum_{\mathbf{e}\in\mathcal{E}}\lVert\mathcal{I}_{e}(x).\exp({n_e\tau}) -   \mathcal{I}_{vr}(\mathbf{e})\rVert,
\label{eq:eventLoss}
\end{equation}
where $\tau$ is the intensity threshold for events to trigger.  Note that $\tau = \Delta L/n_e$, for logarithmic change in radiance $\Delta L$.  It goes without saying that when the events are only monochromatic, the above loss is computed after accordingly converting the 3-dimensional colors to monochrome.

\subsection{Sampling of Events}
While using the loss function derived in~\eqref{eq:eventLoss}, we employ two strategies for event sampling namely, (i) void and (ii) active. The former aims for better visual consistency whereas the latter improves computational efficiency.

\paragraph{Void sampling.}
For some arbitrary time stamp $t$, we randomly select a 2D location $x$ where no event takes place since the last RGB image is acquired. It is intuitive that the color changes minimally for these void events. To impose this constraint, we sample 5\% void events and set their effective event count $n_e=0$. We augment this set of void events to $\mathcal{E}$, while computing the loss of~\eqref{eq:eventLoss}.

\begin{figure*}
\begin{center}
\includegraphics[width=1.0\linewidth]{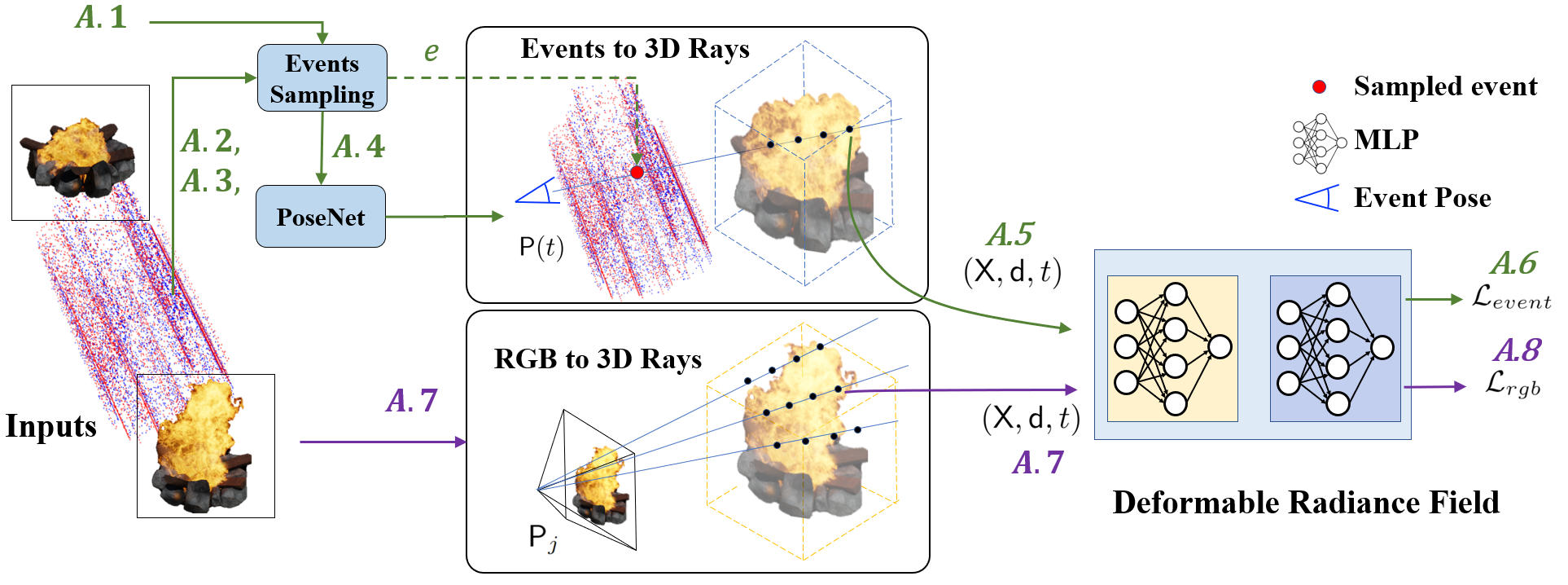}
\end{center}
  \caption{Overview of the proposed method. Notations $A.n$ can be referred to step $n$ accordingly in Algorithm~\ref{alg:algo}.}
\label{fig:diagram2}
\end{figure*}

\paragraph{Active sampling.} 
In the case of the rigid scene and moving camera, it is apparent that the events may not play a significant role in our setup. Instead, they merely introduce the computation burden. The same can be said for the rigid or mostly-rigid parts of the non-rigid scenes. Therefore, we prioritize using events that are generated from the deformable parts. However, such knowledge is not available to us. Therefore, we actively select the desired events during learning. During this, we follow two steps: (i) For $t_j\in\mathcal{T}_s$, we occasionally render the magnitude of the deformation $\omega(\mathsf{X},t_j)$ from $\mathsf{P}_j\in\mathcal{R}$. (ii) The probability of sampling events near $\mathcal{I}_j$ is then set directly proportional to the rendered deformation magnitude at the event's location.

\section{Method Overview}
In this section, we present the complete pipeline of our method, as shown in Figure~\ref{fig:diagram2}. As can be seen, the events are continuously recorded whereas the images are only sparse along the temporal dimension. These sparse images help us to capture the global structure of the radiance field. The finer structures, both in space and time, are then enhanced by using the events. These two aspects however are optimized jointly in an end-to-end manner. 

\subsection{RGB Cameras for Deformable Fields} We supervise the  deformable implicit neural radiance field, $\phi_\theta:(\mathsf{X},\mathsf{d},t)\rightarrow (\mathsf{c},\sigma)$  also using the photometric loss for RGB cameras. When the time-stamp of the calibrated camera is given, the photometric loss is rather straightforward. Let $\mathcal{I}_{vr}(\mathsf{P}_j,t_j)$ be the RGB image rendered from pose $\mathsf{P}_j$ for time $t_j\in\mathcal{T}_j$, the photometric loss for RGB cameras is given by,
\begin{equation}
\mathcal{L}_{rgb} = \sum_{\mathcal{I}_j\in\mathcal{R}}\lVert\mathcal{I}_{j} -   \mathcal{I}_{vr}(\mathsf{P}_j,t_j)\rVert.
\label{eq:rgbLoss}
\end{equation}
Note that the rendered image is the function of both pose and time-stamp of the corresponding image $\mathcal{I}_j$, as the radiance field is temporally deforming. The above loss function supervises the deforming field only sparsely in time, from the RGB images' poses at the corresponding acquisition time.

\subsection{Sparse Poses for Dense Events}
Recall that the sparse pairs of camera pose and time $(\mathsf{P}_j,t_j)$ for $t_j\in\mathcal{T}_s$ are available in our setting.  We use this information in two ways: (i) each $\mathsf{P}_j$ is directly used to cast the rays and render the image required for $\mathcal{L}_{rgb}$ computation in~\eqref{eq:rgbLoss}; (ii) the sparse pairs $(\mathsf{P}_j,t_j)$ 
are used in the time-to-pose mapping network \textbf{PoseNet}. Instead of predicting the pose directly from time, we predict the residual pose, where the initial pose for a given event is obtained by local temporal interpolation of available RGB poses.

\subsection{The Algorithm}
We summarize the loss computation of our method in Algorithm~\ref{alg:algo}. Using the derived loss, three multilayer perceptrons, each for PoseNet, deformation field, and radiance in the canonical frame, are trained. Further implementation details of our method are presented in the next section.  

\begin{algorithm}
{
\small
\caption{\small $\mathcal{L}_{total}$ = computeTotalLosse($\mathcal{R},\mathcal{E},\lambda)$}
\label{alg:algo}
\begin{algorithmic}
\STATE 1. Render the warp field $\omega(\mathsf{X},t)$ at time $t_j$ from nearby $\mathsf{P}_j$. 
\STATE 2. Sample active events $\mathcal{E}_a\subset\mathcal{E}$ using the rendered wrap field. 
\STATE 3. Sample void events $\mathcal{E}_v$ and set  $\mathcal{E}_{total} = \mathcal{E}_a\cup \mathcal{E}_v$.
\STATE 4. For each event $\mathbf{e}\in\mathcal{E}_{total}$, obtain the pose using \textbf{PoseNet}.
 \STATE 5. Render event ray $\mathcal{I}_{vr}(\mathbf{e})$ with event pose and pixel location.
 \STATE 6. Compute event loss $\mathcal{L}_{event}$ for $\mathcal{E}_{total}$ using~\eqref{eq:eventLoss}.
  \STATE 7. Sample each ray from $\mathcal{I}_j\in\mathcal{R}$ and render  $\mathcal{I}_{vr}(\mathsf{P}_j,t_j)$.
\STATE 8. Compute the photometric loss $\mathcal{L}_{rgb}$ using~\eqref{eq:rgbLoss}.

 \STATE 9. Return $\mathcal{L}_{total}=\mathcal{L}_{event}+\lambda\mathcal{L}_{rgb}$ 
\end{algorithmic}
}
\end{algorithm}

\section{Implementation Details}

For the Pose correction network (PoseNet), we use two 8-layer MLPs with the hidden size 256 to learn the translation and rotation residuals. We initialize translation and rotation using cubic spherical linear interpolations, respectively. We use a 6-layer MLPs with a width 128 for the deformation network and output 8 dimension latent codes. We utilize coarse-to-fine regularization to modulate the positional encoding components, as suggested in \cite{park2021nerfies}. We train on 4 NVIDIA GeForce RTX 2080 Ti using 64 coarse rays and 128 fine ray samples. $\lambda$=10 is used for RGB loss and in supplementary Tab. 3 we show sensitivity analysis on different $\lambda$.%

\paragraph{Synthetic Data.}
Due to the absence of publicly available benchmarks or relevant synthetic event stream datasets and the insufficient number of monocular frames provided by works such as \cite{park2021nerfies} \cite{park2021hypernerf}\cite{mildenhall2021nerf} to simulate events, we create our own datasets with varying degrees and types of motion using Blender and simulate events using ESIM\cite{rebecq2018esim}. 
We synthesize 3 different scenes. \textbf{Non-rigid Lego}: We created a 360-degree camera path around the Lego with the moving `blade' for two cycles of upward and downward motion which is 4 times faster than \cite{pumarola2021d}. The challenge posed by this dataset is to determine whether the model is capable of effectively acquiring knowledge on the locally-rigid transformation of the blade, and discerning it from the stationary component. \textbf{Campfire}: In contrast to Lego, the majority part of the burning campfire dataset is dynamic. The challenge lies in the ability to learn the variations of flame contours and colour from highly varying unordered boundaries. \textbf{Fluid}: Water flowing out from pipes hitting the ground with a lot of splashes. The reconstruction of water flow presents the most challenging data, due to its intricate shape variations, colour variations, and the effects of light and shadows. To produce high-temporal resolution events we use Blender to render thousands of continuous frames. In contrast to the approach in \cite{park2021nerfies}, which employs a prior based on static 3D background points for regularization, we configure the synthetic dataset's background as white, thereby utilizing it as a means to regularize the background during the training process of the radiance field\cite{mildenhall2021nerf}. All poses are accurate for all synthetic datasets, and the $\tau$ of events is also recorded.

\paragraph{Real Data.} We evaluate our method on the public datasets which contain dynamic scenes. 
a) HS-ERGB: We evaluate our method to model deformation on High-Speed Events and RGB dataset \cite{Tulyakov21CVPR} which include challenging dynamic scenes such as a rotating \textbf{Umbrella} as well as the \textbf{Candle} and \textbf{Fountain}. Note that in this dataset the camera is static so we disable the PoseNet for residual learning. This dataset provides high-resolution event stream and RGB images. The extrinsics between RGB and event cameras, as well as the pixel-wise alignments, are also provided. 
b) CED: To evaluate our method on the human subject we use the dynamic \textbf{Selfie} sequence in Color Event Camera Dataset \cite{scheerlinck2019ced} which contains both colour frames and colour events from the DAVIS 346C. 
c) EVIMOv2: We choose the EVIMOv2 Dataset to evaluate our method on the dynamic scenes with moving cameras. The dataset\cite{burner2022evimo2} provides millimeter-accurate object poses from a Vicon motion capture system. We use the Samsung DVS Gen3 camera with Flea3 (RGB) as they share most of the field of view. We downsample the rgb frame from 2080×1552 to half and align the frame with events using the depth provided by the Vicon pose estimate and 3D scanning. We selected two sequences, namely the \textbf{Toycar} with a moderate moving speed and the \textbf{Quadcopter} with a high motion speed. 

Whenever the contrast threshold $\tau$ is not available (or unreliable) for real data,  we estimate per-pixel positive and negative thresholds by comparing nearby RGB images and the intermediate event counts~\cite{brandli2014real}. 
We also filter out mismatched events during this process. For all sequences, we subsample the original high-speed video for training and use the intermediate frames for validation.

\begin{table*}
\resizebox{\linewidth}{!}{
\begin{tabular}{@{}lcccccccccccc@{}}\toprule
& \multicolumn{4}{c}{\textbf{Lego}} & \multicolumn{4}{c}{\textbf{Campfire}}  & \multicolumn{4}{c}{\textbf{Fluid}}\\

\cmidrule(r){2-5} \cmidrule(r){6-9} \cmidrule(r){10-13}
{Methods}& MSE $\downarrow$ & PSNR $\uparrow$ & SSIM $\uparrow$& LPIPS $\downarrow$& MSE$\downarrow$ &  PSNR$\uparrow$ & SSIM$\uparrow$ & LPIPS $\downarrow$ & MSE $\downarrow$ & PSNR$\uparrow$& SSIM$\uparrow$ &  LPIPS $\downarrow$ \\ 
\midrule
NeRF~\cite{mildenhall2021nerf} & 5.54 & 22.11 & 0.89 & 0.229 &  7.11 &  21.01 & 0.85 & 0.206 & 3.52 & 24.94 & 0.83 & 0.324 \\
HyperNeRF~\cite{park2021hypernerf} & 4.40 & 24.28 & 0.94 & 0.080 & 5.27 & 22.94 & 0.93 & 0.152 & 3.14 & 25.17 & 0.85  &0.303\\
Nerfies~\cite{park2021nerfies} & 2.90 & 25.97 & 0.96 & 0.089 & 5.13 & 23.11 & 0.92 &0.154 & 2.47 & 25.25 & 0.87 & 0.300\\
DE$-$Baseline & 2.10 & 27.12 & 0.97 & 0.093 &  4.45 &  23.76 &  0.93  & 0.143 & 2.51 & 26.07 & 0.87 & 0.296\\

\midrule
DE$-$Nerf(Ours) & \textbf{0.32} & \textbf{35.04} & \textbf{0.99} & \textbf{0.034} & \textbf{1.95} & \textbf{27.56}  & \textbf{0.96} & \textbf{0.115} & \textbf{1.91} & \textbf{26.92} & \textbf{0.91} & \textbf{0.289}  \\
\bottomrule
\end{tabular}}
\vspace{1mm}
\caption{ Comparison of our method against the state-of-the-art and the established baseline, on realistically rendered artificial scenes. } 
\label{tab:syt_table}
\end{table*}


\begin{table*}

\resizebox{\linewidth}{!}{
\begin{tabular}{@{}lcccccccccccc@{}}\toprule
(Static-cameras) & \multicolumn{4}{c}{\textbf{Umbrella}} & \multicolumn{4}{c}{\textbf{Candle}} & \multicolumn{4}{c}{\textbf{Fountain}}\\
\cmidrule(r){1-1}  \cmidrule(r){2-5} \cmidrule(r){6-9} \cmidrule(r){10-13}
{Methods}& MSE $\downarrow$ & PSNR $\uparrow$ & SSIM $\uparrow$& LPIPS $\downarrow$& MSE$\downarrow$ &  PSNR$\uparrow$ & SSIM$\uparrow$ & LPIPS $\downarrow$ & MSE $\downarrow$ & PSNR$\uparrow$& SSIM$\uparrow$ &  LPIPS $\downarrow$ \\ 
\midrule
NeRF~\cite{mildenhall2021nerf} & 2.72 & 25.41  & 0.81  & 0.471 & 11.0 & 19.42 & 0.86 & 0.333 & 6.54 & 21.89 & 0.48 & 0.600  \\
HyperNeRF~\cite{park2021hypernerf} & 2.14 & 26.72 & 0.85 & 0.432 & 4.01 & 27.04 & 0.94 & 0.283 & 5.31 & 22.80 & 0.52 & 0.688  \\
Nerfies~\cite{park2021nerfies} & 1.77 & 28.30 & 0.89 & 0.358 & 4.23 & 26.08 & 0.93 & 0.293 & 3.95 & 24.06 & 0.66 & 0.552 \\
DE$-$Baseline & 2.04 & 27.19 & 0.86 & 0.432 & 4.75 & 25.72 & 0.93 & 0.246 & 4.13 & 23.87 & 0.55 & 0.610 \\

\midrule
DE$-$NeRF (Ours) & \textbf{0.45} & \textbf{33.44} & \textbf{0.95} & \textbf{0.341} & \textbf{0.38} & \textbf{34.22} & \textbf{0.97} & \textbf{0.242} & \textbf{3.11} & \textbf{25.13} & \textbf{0.71} & \textbf{0.546} \\
\midrule
(Moving-cameras) & \multicolumn{4}{c}{\textbf{Selfie}} &  \multicolumn{4}{c}{\textbf{Toycar}} & \multicolumn{4}{c}{\textbf{Quadcopter}}\\
\cmidrule(r){1-1}  \cmidrule(r){2-5} \cmidrule(r){6-9} \cmidrule(r){10-13}
{Methods}& MSE $\downarrow$ & PSNR $\uparrow$ & SSIM $\uparrow$& LPIPS $\downarrow$& MSE$\downarrow$ &  PSNR$\uparrow$ & SSIM$\uparrow$ & LPIPS $\downarrow$ & MSE $\downarrow$ & PSNR$\uparrow$& SSIM$\uparrow$ &  LPIPS $\downarrow$ \\ 
NeRF~\cite{mildenhall2021nerf} & 6.25 & 22.41  & 0.83  & 0.382 & 3.93 & 24.17 & 0.85 & 0.406 & 7.49 & 21.25 & 0.77 & 0.553 \\
HyperNeRF~\cite{park2021hypernerf} & 3.76 & 25.02 & 0.90 &0.334 & 1.41 & 31.45 & 0.94 & 0.242& 2.66 & 27.62 & 0.92 & 0.263\\
Nerfies~\cite{park2021nerfies} & 2.77 & 25.85 & 0.91 & 0.303 & 0.87 & 33.09 & 0.96 & 0.217 & 2.25 & 28.69 & 0.93 & 0.244 \\
DE$-$Baseline & 3.79 & 24.39 & 0.89 & 0.396 & 1.14 & 31.99 & 0.95 & 0.223 & 3.01 & 27.54& 0.90 & 0.265\\

\midrule
DE$-$NeRF (Ours) &\textbf{1.80}  & \textbf{27.74} & \textbf{0.94} & \textbf{0.224} & \textbf{0.54} & \textbf{34.17} & \textbf{0.98} & \textbf{0.201} & \textbf{1.53} & \textbf{29.95} & \textbf{0.95}  & \textbf{0.210}\\
\bottomrule
\end{tabular}}
\vspace{1mm}
\caption{Real data experiments on two cases: static camera with dynamic scene (top); and moving camera with dynamic scene (bottom). In all six real-world diverse datasets our method performs significantly better than the state-of-the-art methods and established baseline.}
\label{tab:real_table}
\end{table*}


\paragraph{{Baselines}.}
Since there exists no method that exploits event cameras to model deformable radiance fields, we establish a new baseline method -- which we refer to as \textbf{DE-baseline}. This baseline is inspired by two notable works on deformable NeRF~\cite{park2021nerfies} and event-based NeRF~\cite{rudnev2022eventnerf}.  For DE-Baseline, we sample one ray each for two neighbouring events.
Sampled rays are passed through the deformable radiance field, as in our method, using the exact same network. 
Then, we compute the event loss proposed in~\cite{klenk2023nerf}, together with the photometric loss of~\eqref{eq:rgbLoss}, for supervision. Similarly to \cite{klenk2023nerf} we use the normalized brightness increments loss \cite{hidalgo2022event} for real-world cases. In our real data experiments, we found that the normalized event loss is detrimental to PSNR. This aligns with the observation of~\cite{klenk2023nerf}, which can largely be attributed to noisy events.   

We also compare our method against two state-of-the-art methods, namely, \textbf{Nerfies} \cite{park2021nerfies} and \textbf{HyperNeRF} \cite{park2021hypernerf}, that aim to model the deformable scenes in RGB-only settings. In order to highlight the difficulty, we report the results with the well-known rigid \textbf{NeRF} method~\cite{mildenhall2021nerf}. Drawing inspiration from other event-based NeRF we also report results using events-to-frame method as reference.



\paragraph{Evaluation Matric.} We evaluate our method in learning high-speed dynamic scenes using the following metrics: (i) MSE (with a factor of $\times 10^{-3}$), (ii) Peak signal-to-noise ratio,  (iii) The structural similarity (SSIM)~\cite{wang2004image}, and (iv) Learned Perceptual Image Patch Similarity (LPIPS)~\cite{zhang2018unreasonable} using VGG. We also follow~\cite{zhu2022nice} to calculate the pose error using ATE-RMSE.

\section{Experiments}
\label{sec:exp}

\begin{figure}
\begin{center}
\includegraphics[width=0.99\linewidth]{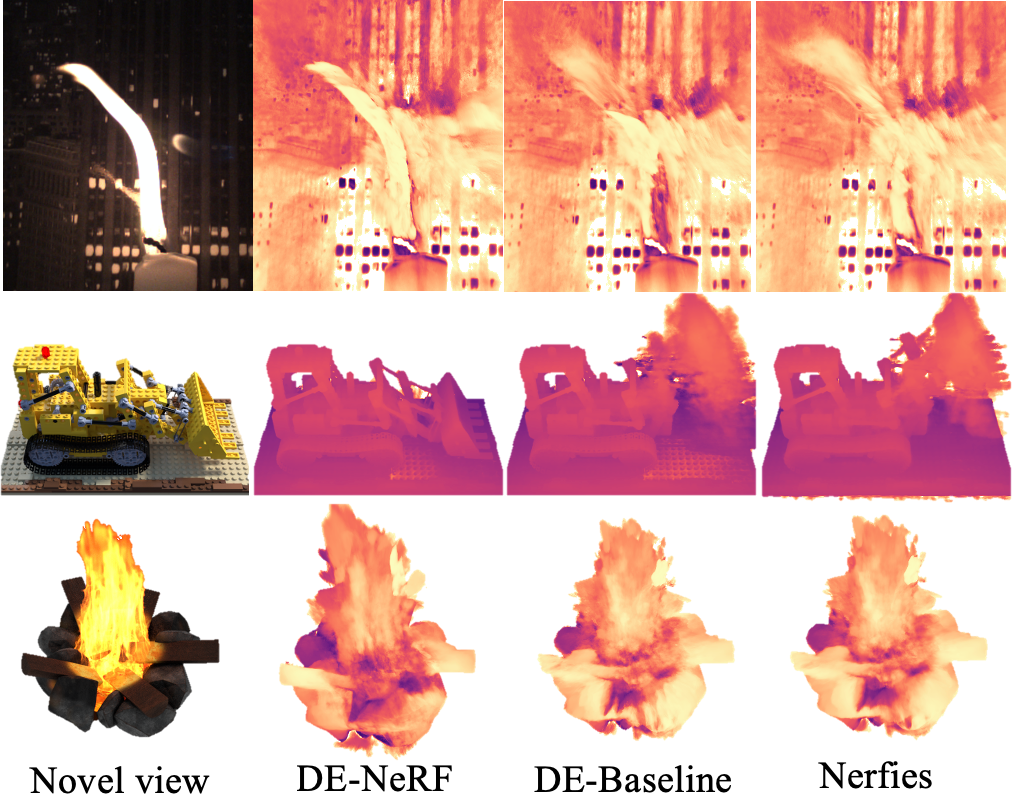}
\end{center}
  \caption{Comparison of rendered depths on three datasets of Tables~\ref{tab:syt_table}\&\ref{tab:real_table}. Our method (DE-NeRF) provides very realistic depths. }
\label{fig:Qualitative_depth}
\end{figure}

\newcolumntype{C}{ >{\centering\arraybackslash} p{0.16\linewidth} }

\begin{table*}
\begin{center}
\begin{tabular}{ @{}rC@{\hskip 3pt}C@{\hskip 3pt}C@{\hskip 3pt}C@{\hskip 3pt}C@{\hskip 3pt}C@{} }
\vspace{-0.5cm}
\rotatebox{90}{Lego} &\includegraphics[width=\linewidth,valign=m]{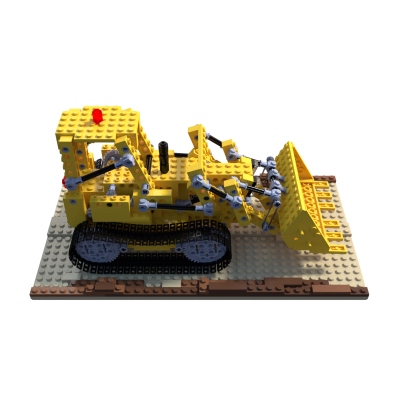} & \includegraphics[width=\linewidth,valign=m]{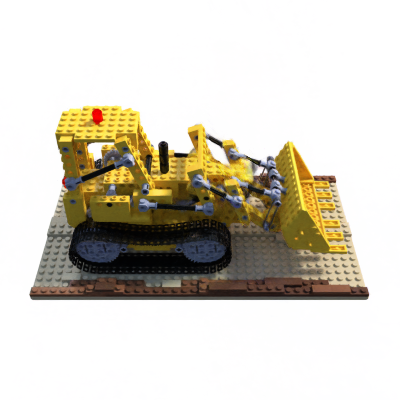}  & \includegraphics[width=\linewidth,valign=m]{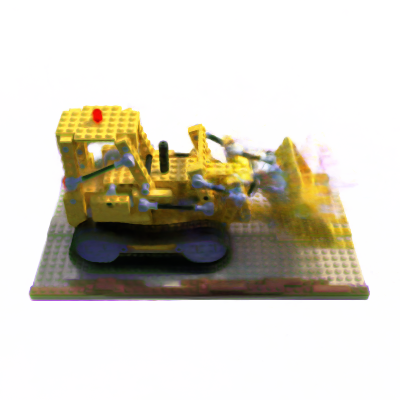} & \includegraphics[width=\linewidth,valign=m]{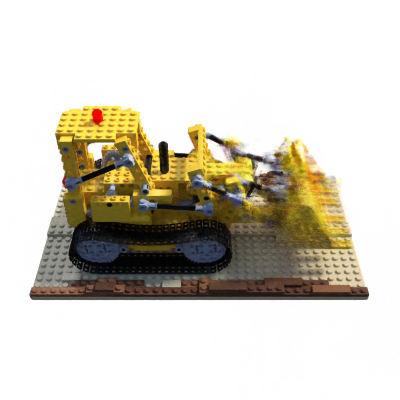} & \includegraphics[width=\linewidth,valign=m]{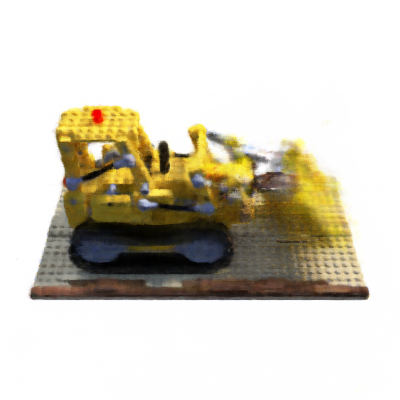} &\includegraphics[width=\linewidth,valign=m]{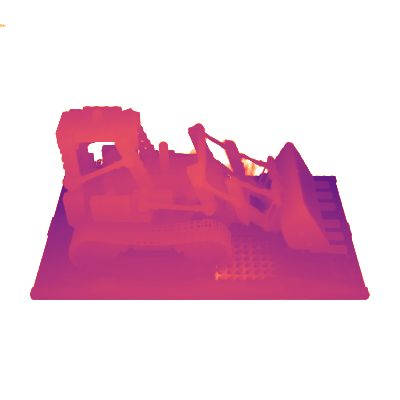 } \\
\vspace{-0.2cm}
\rotatebox{90}{Campfire} & \includegraphics[width=\linewidth,valign=m]{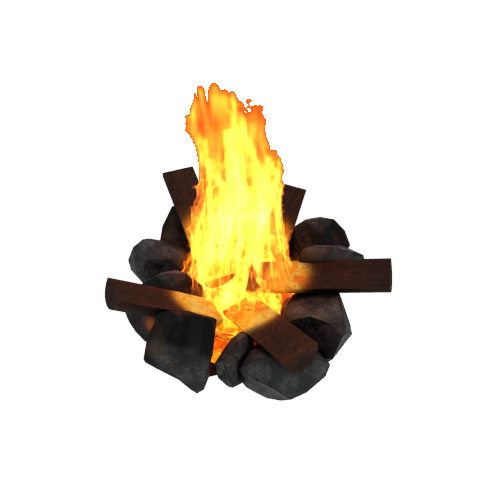} & \includegraphics[width=\linewidth,valign=m]{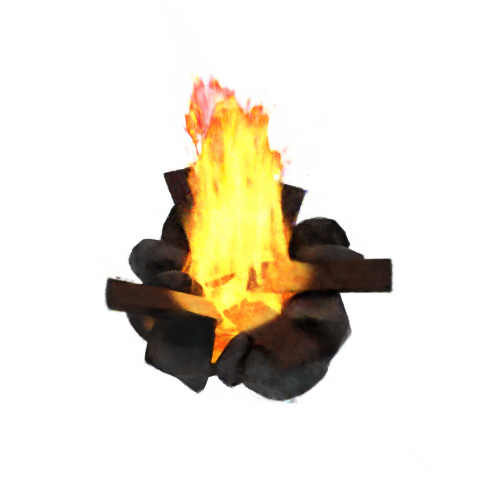} & \includegraphics[width=\linewidth,valign=m]{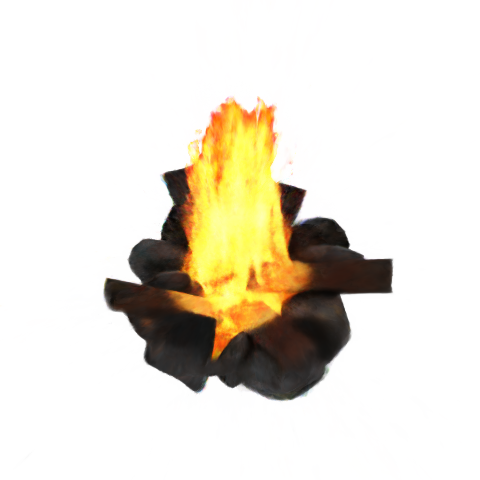} & \includegraphics[width=\linewidth,valign=m]{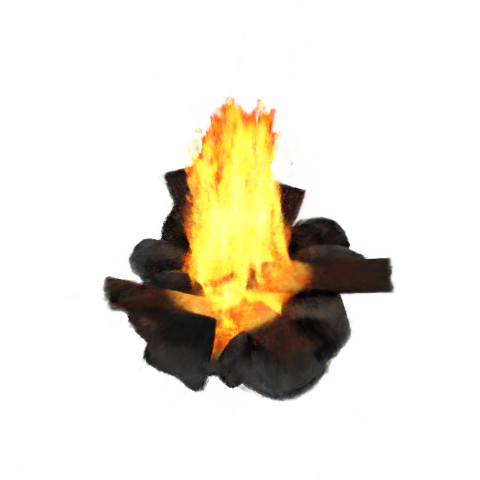} & \includegraphics[width=\linewidth,valign=m]{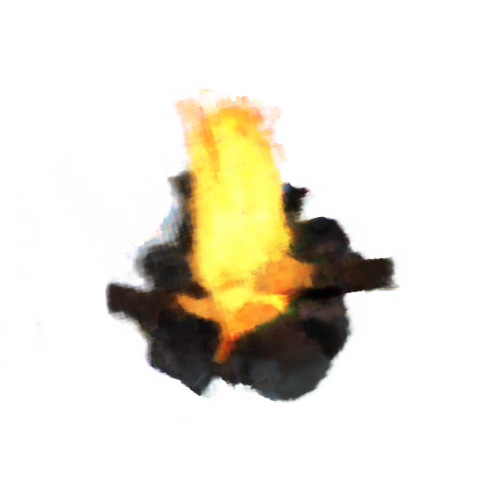} & \includegraphics[width=\linewidth,valign=m]{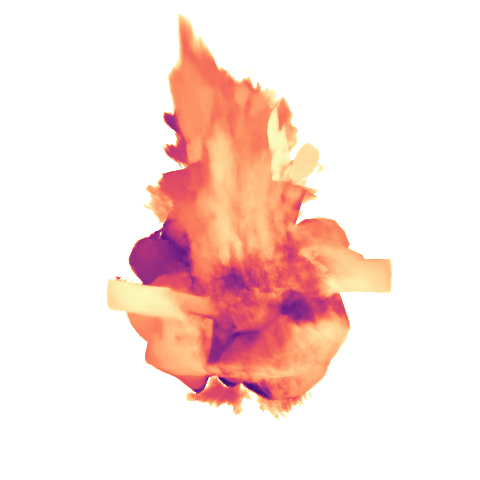 } \\

\vspace{0.1cm}

\rotatebox{90}{Fluid} &\includegraphics[width=\linewidth,valign=m]{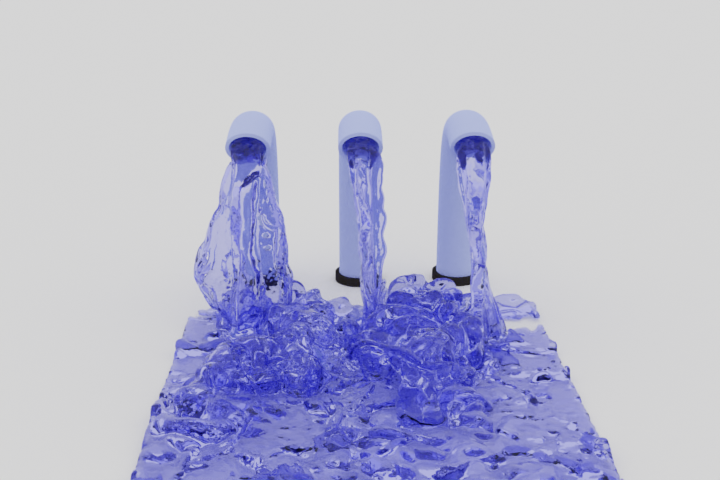} & \includegraphics[width=\linewidth,valign=m]{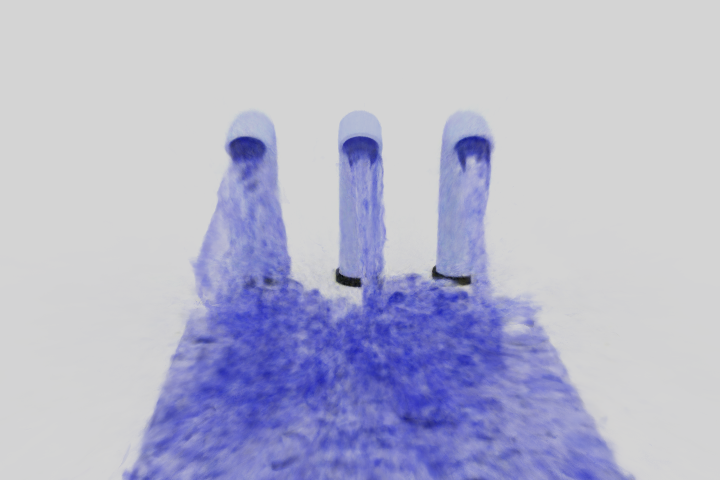} & \includegraphics[width=\linewidth,valign=m]{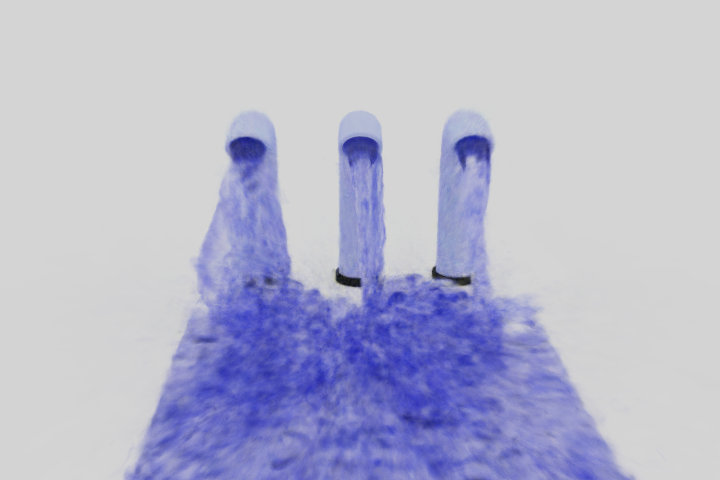} & \includegraphics[width=\linewidth,valign=m]{./figures/fluid/rgb_000671_ours.png} & \includegraphics[width=\linewidth,valign=m]{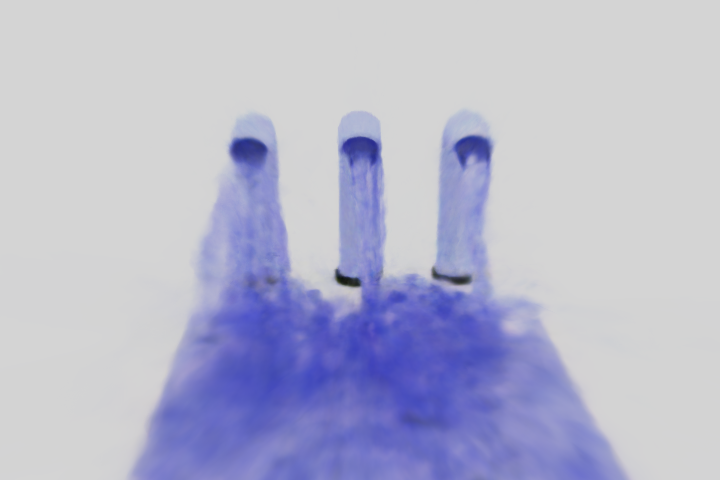} & \includegraphics[width=\linewidth,valign=m]{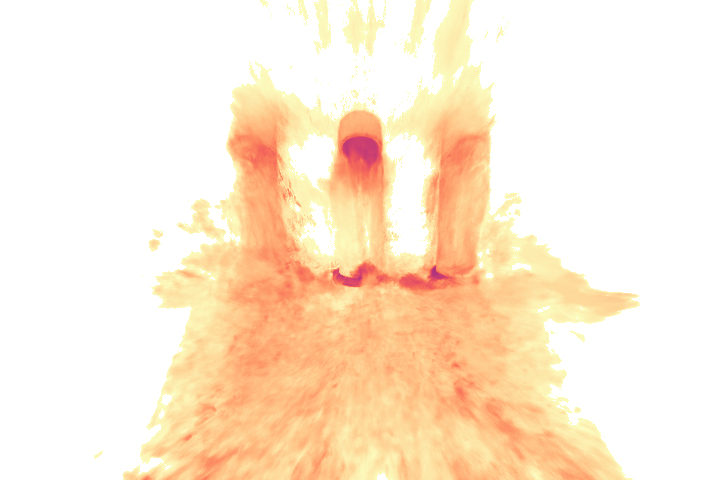 } \\

\vspace{0.1cm}
\rotatebox{90}{Umbrella} & \includegraphics[width=\linewidth,valign=m]{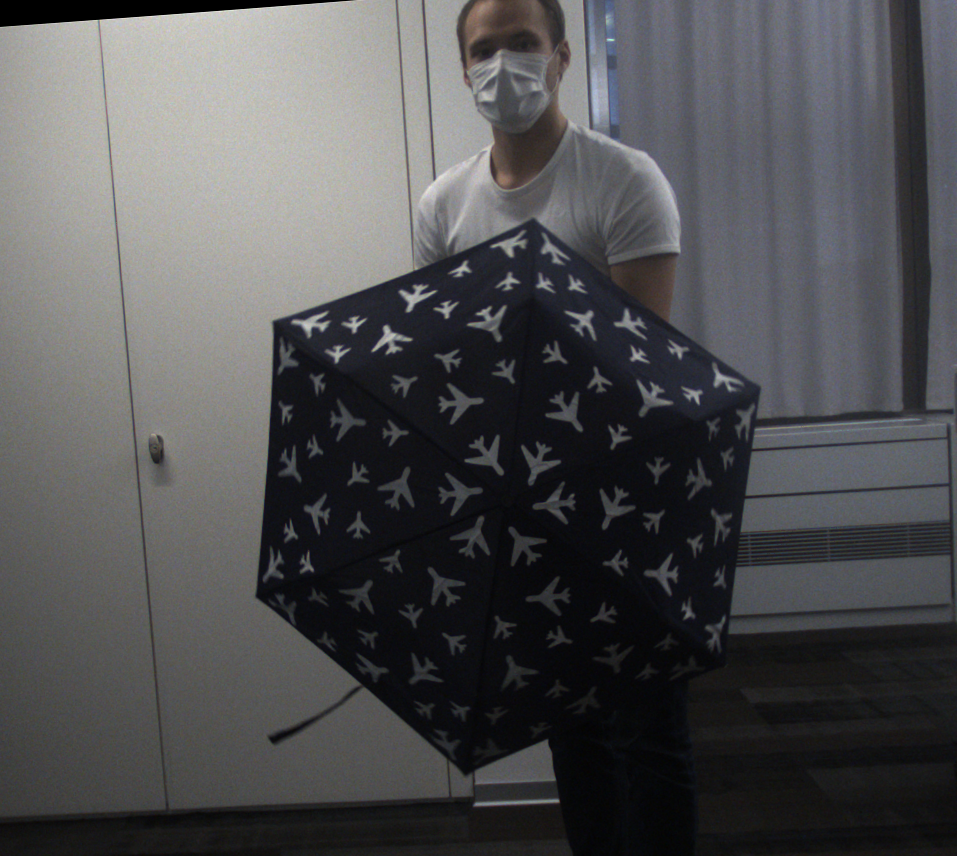} & \includegraphics[width=\linewidth,valign=m]{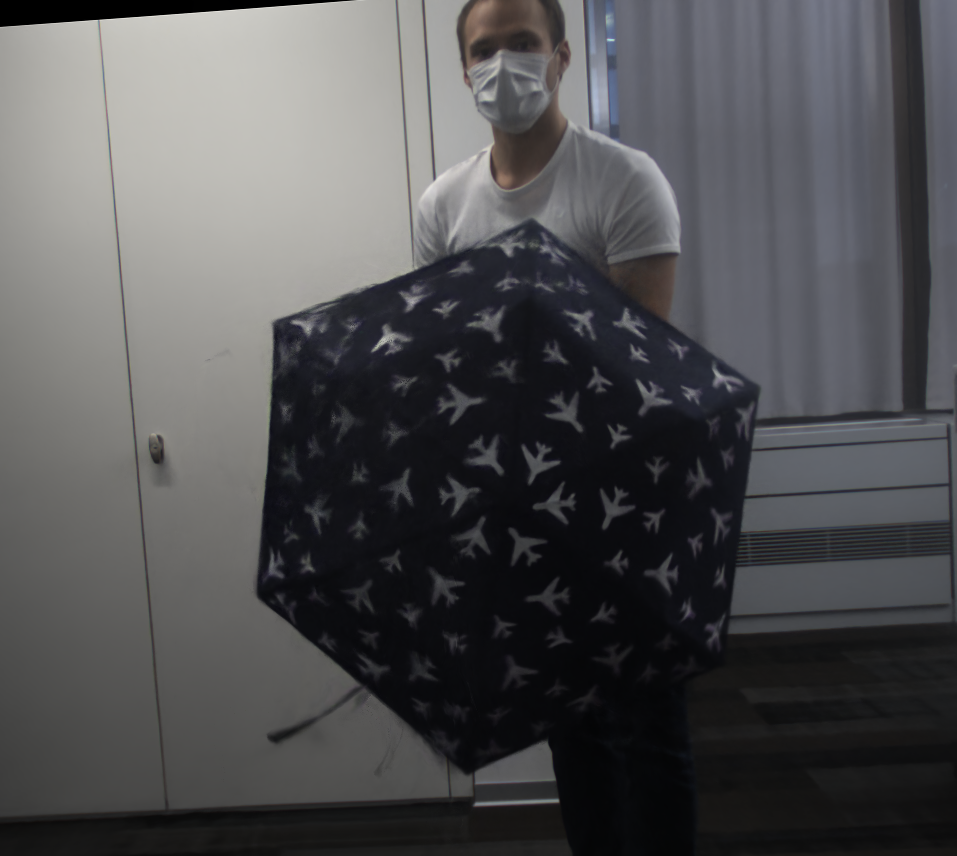} & \includegraphics[width=\linewidth,valign=m]{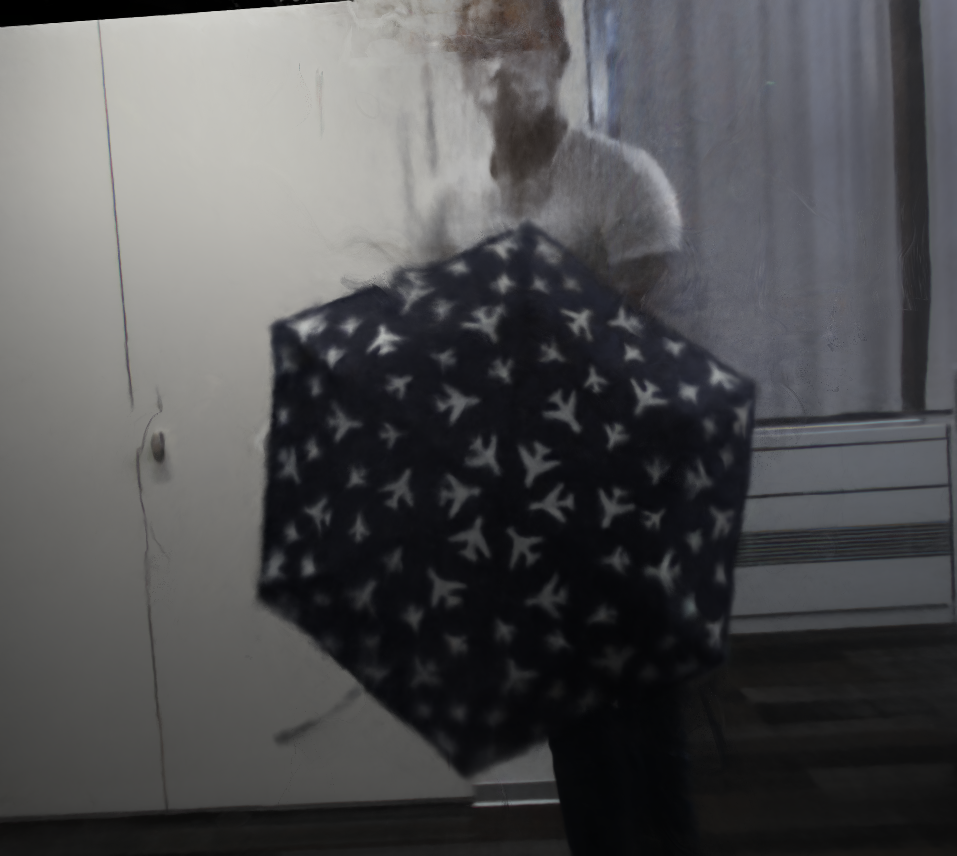} & \includegraphics[width=\linewidth,valign=m]{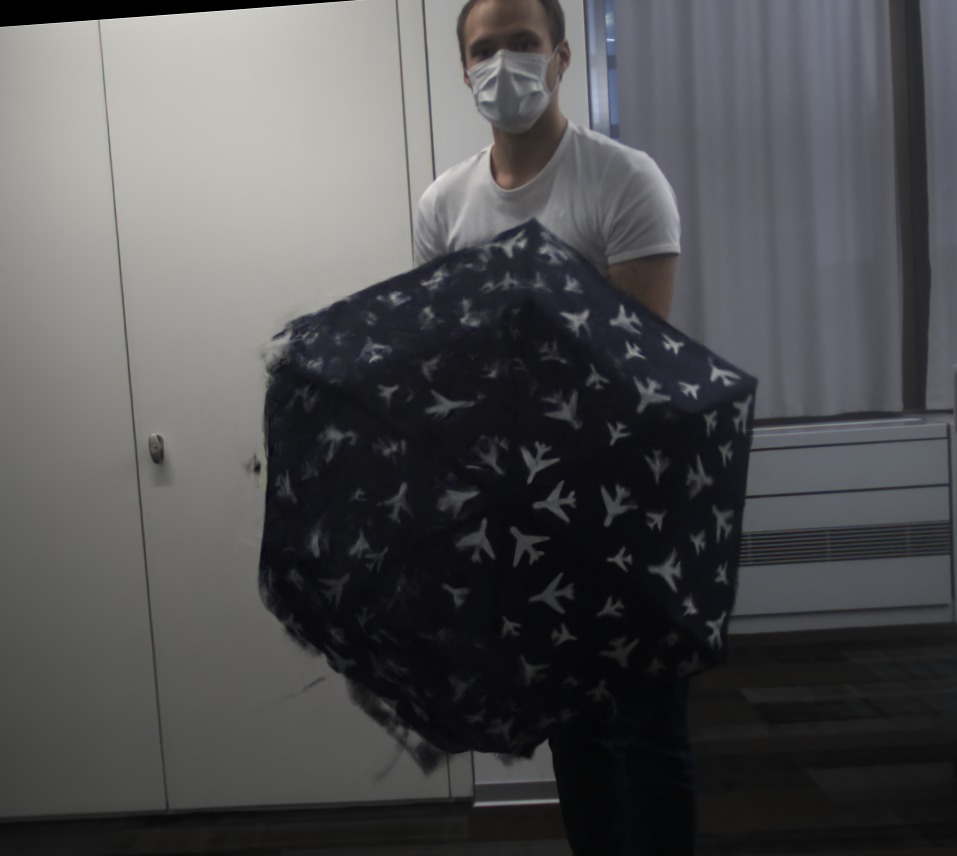} & \includegraphics[width=\linewidth,valign=m]{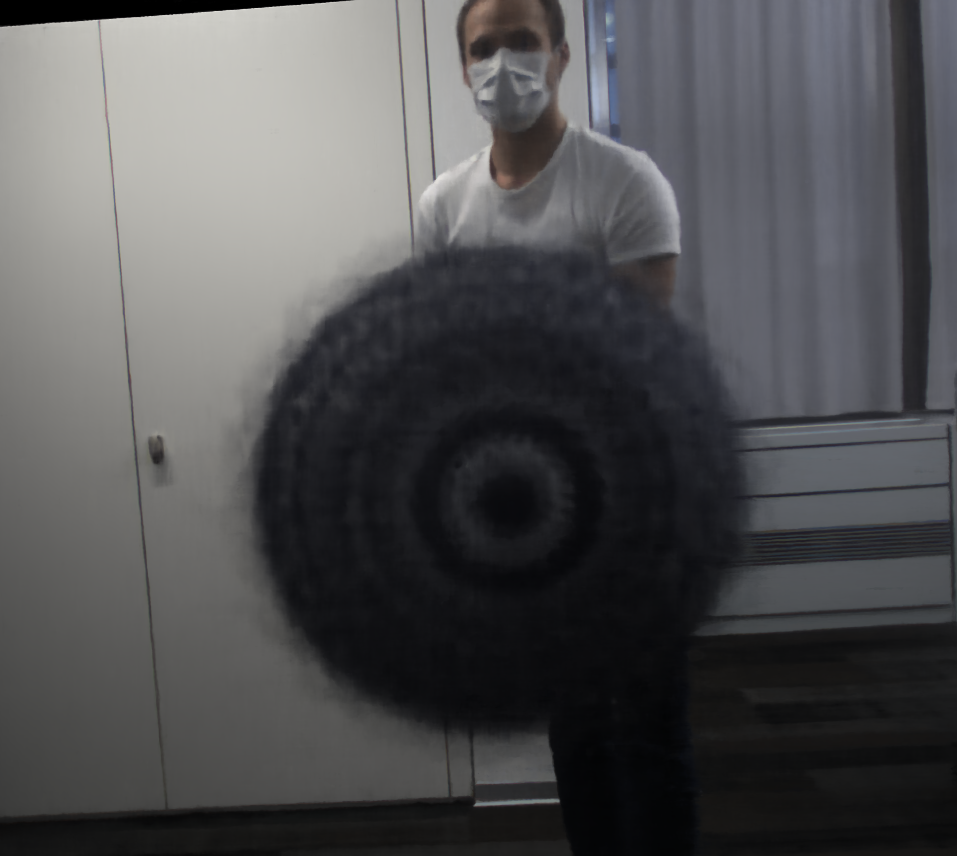} & \includegraphics[width=\linewidth,valign=m]{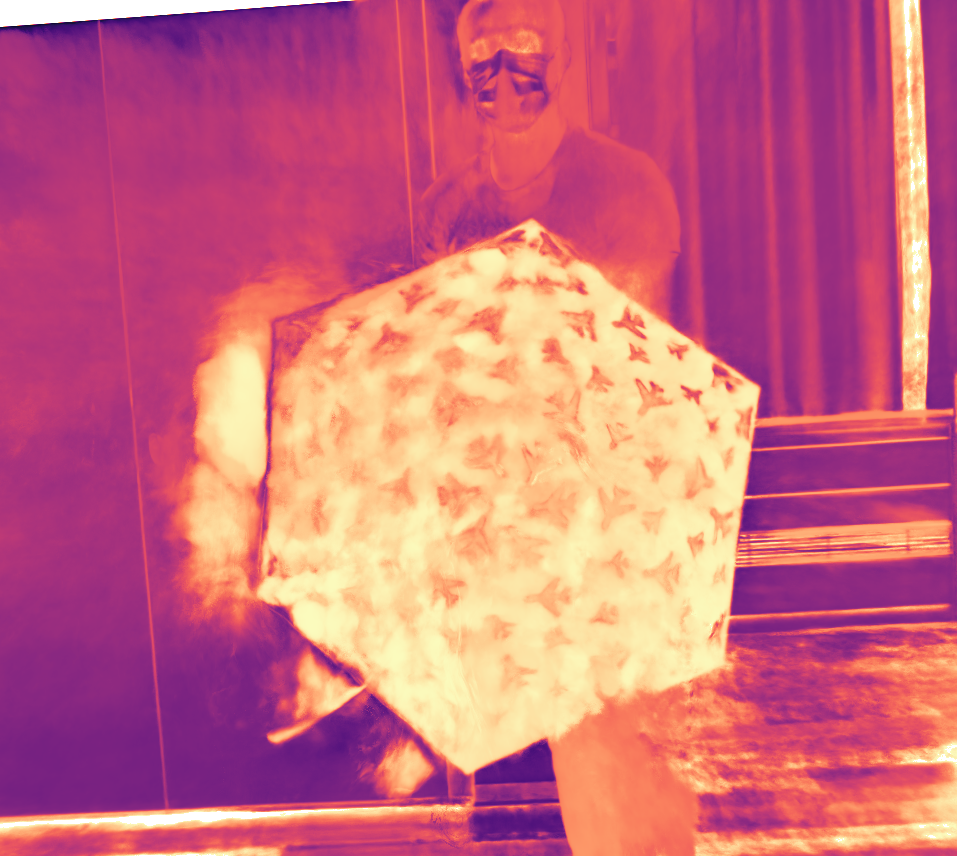} \\

\vspace{0.1cm}
\rotatebox{90}{Candle} & \includegraphics[width=\linewidth,valign=m]{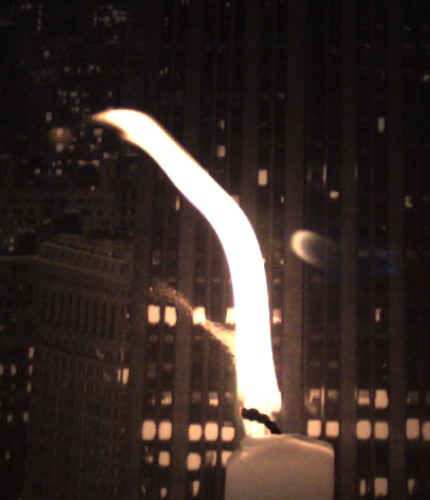} & \includegraphics[width=\linewidth,valign=m]{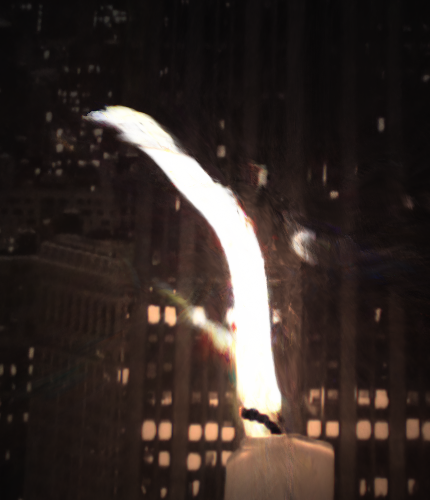} & \includegraphics[width=\linewidth,valign=m]{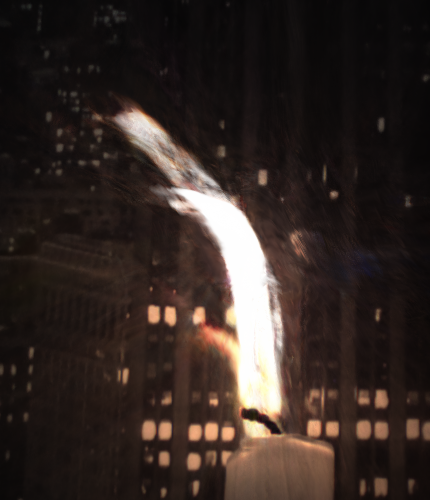} & \includegraphics[width=\linewidth,valign=m]{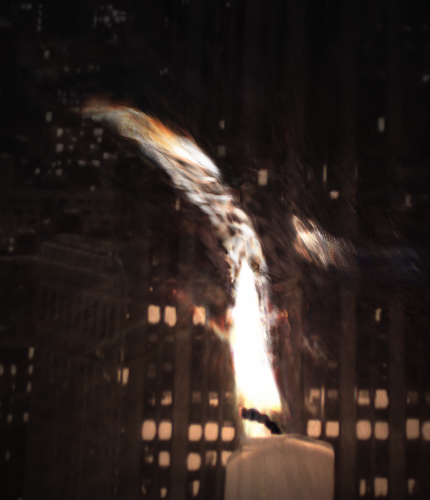} & \includegraphics[width=\linewidth,valign=m]{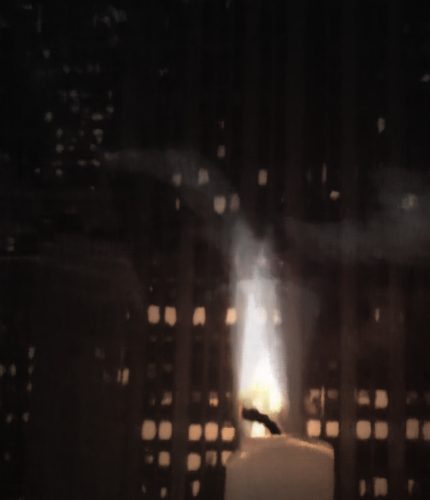} & \includegraphics[width=\linewidth,valign=m]{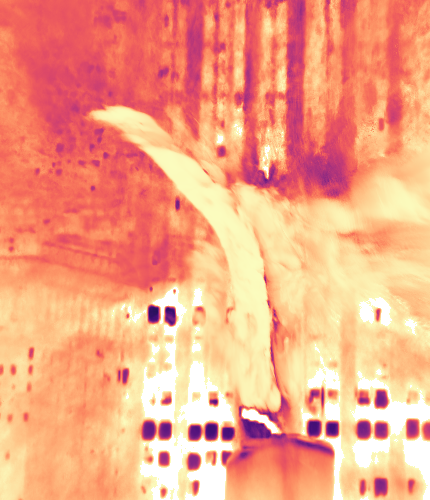} \\
\vspace{0.1cm}
\rotatebox{90}{Fountain} & \includegraphics[width=\linewidth,valign=m]{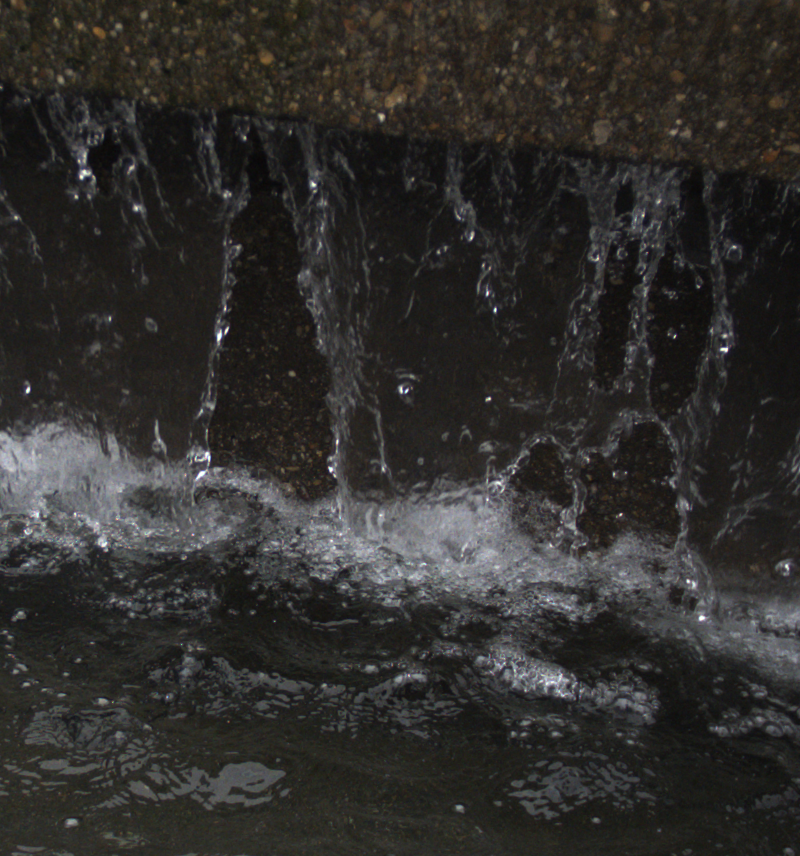} & \includegraphics[width=\linewidth,valign=m]{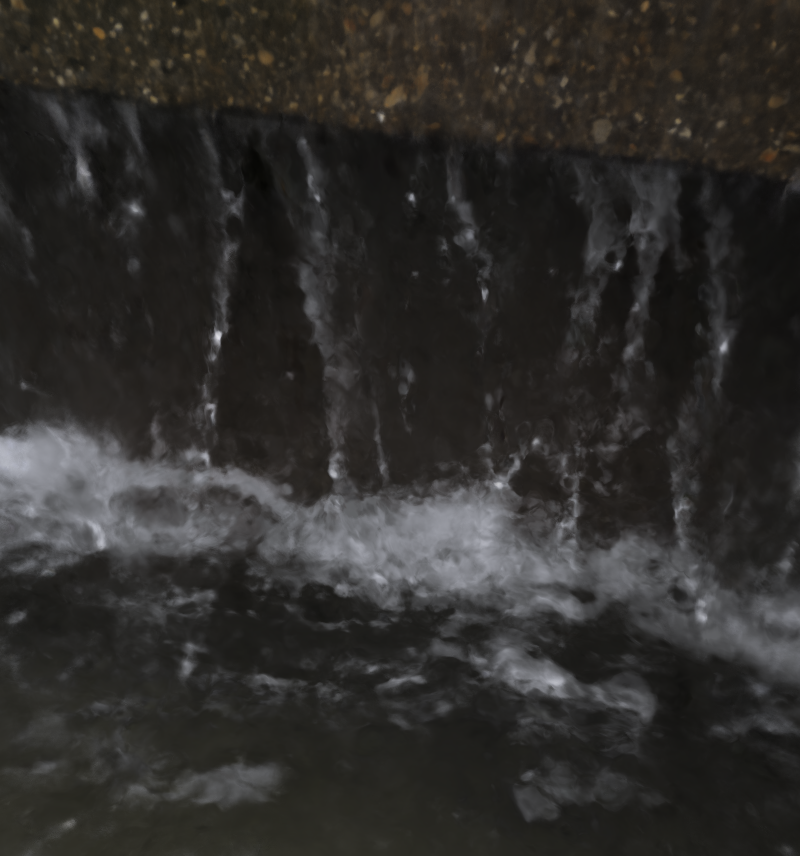} & \includegraphics[width=\linewidth,valign=m]{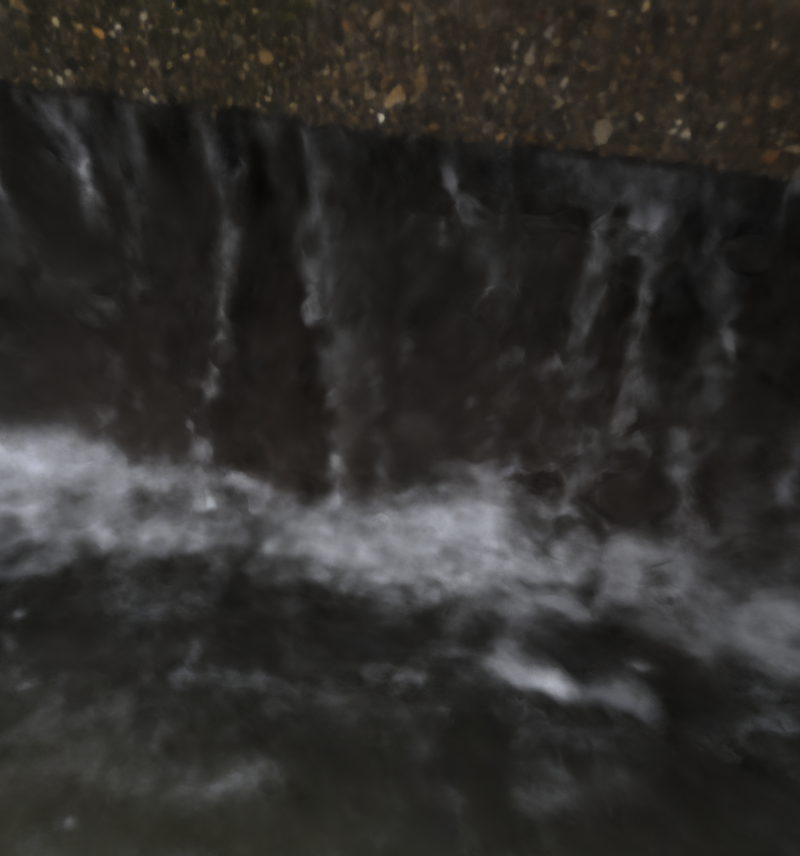} & \includegraphics[width=\linewidth,valign=m]{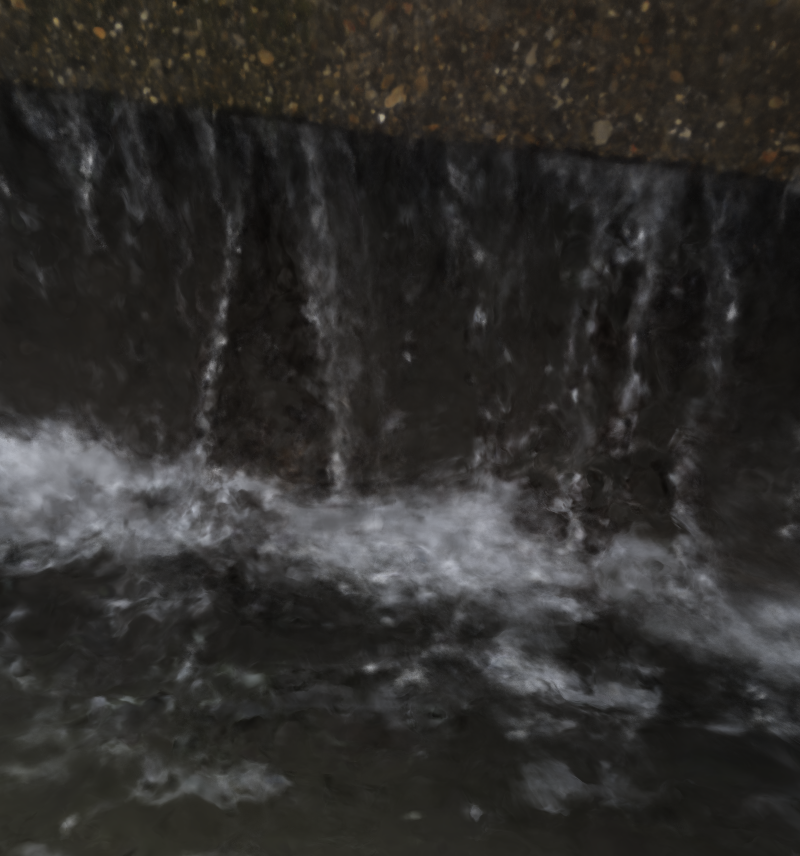} & \includegraphics[width=\linewidth,valign=m]{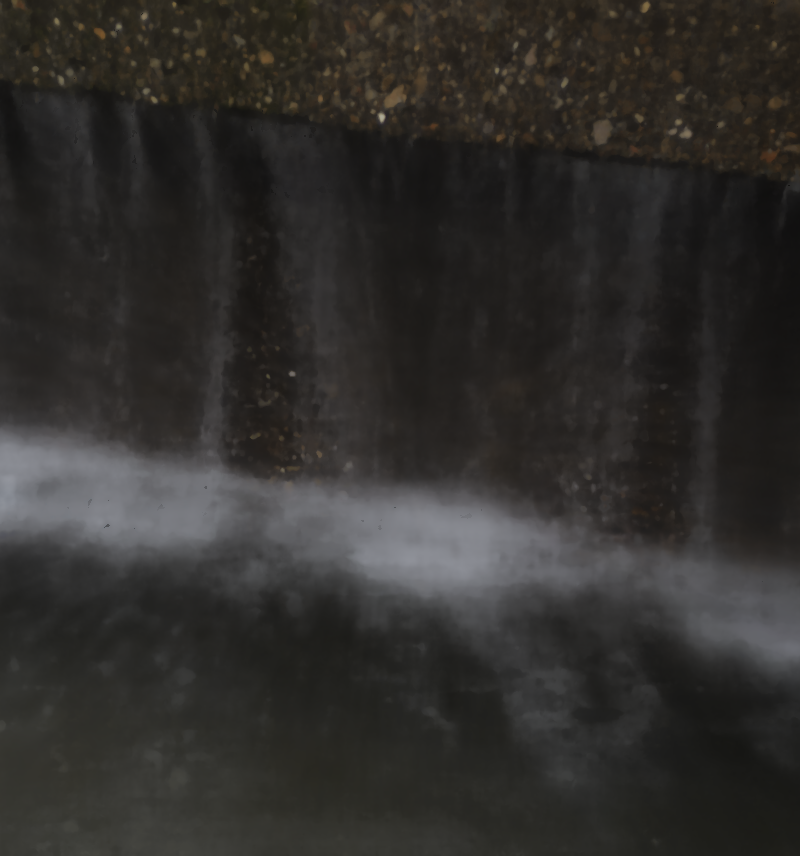} & \includegraphics[width=\linewidth,valign=m]{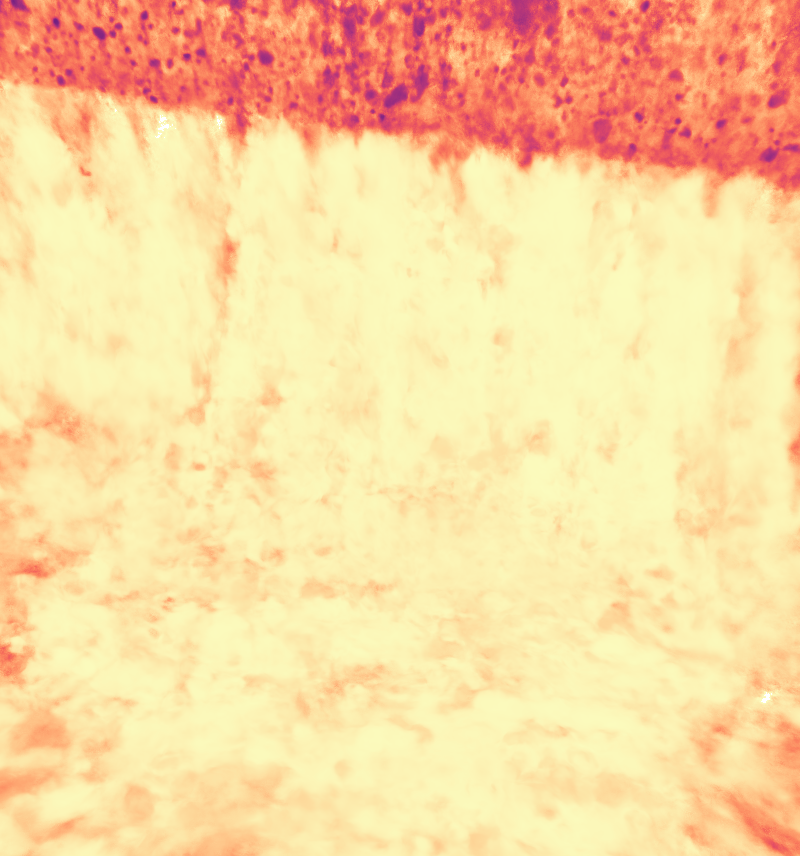 } \\

\vspace{0.1cm}
\rotatebox{90}{Selfie} & \includegraphics[width=\linewidth,valign=m]{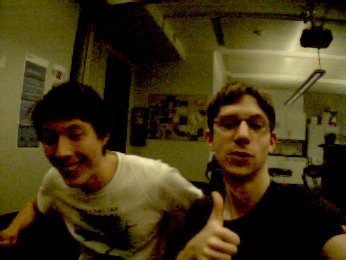} & \includegraphics[width=\linewidth,valign=m]{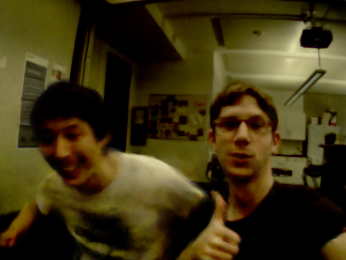} & \includegraphics[width=\linewidth,valign=m]{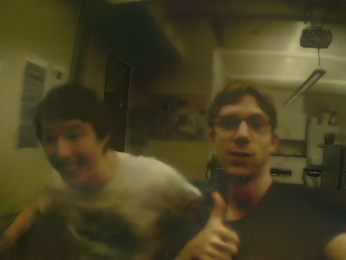} & \includegraphics[width=\linewidth,valign=m]{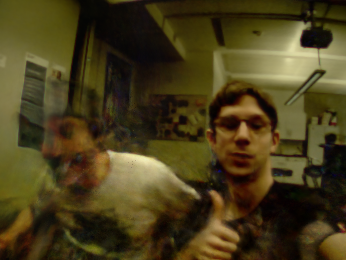} & \includegraphics[width=\linewidth,valign=m]{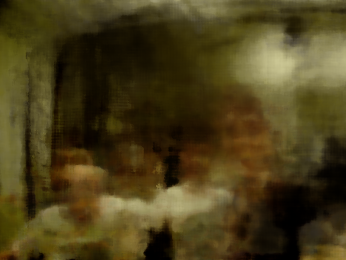} & \includegraphics[width=\linewidth,valign=m]{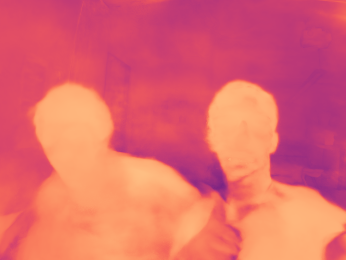} \\

\vspace{0.1cm}
\rotatebox{90}{Toycar} &\includegraphics[width=\linewidth,valign=m]{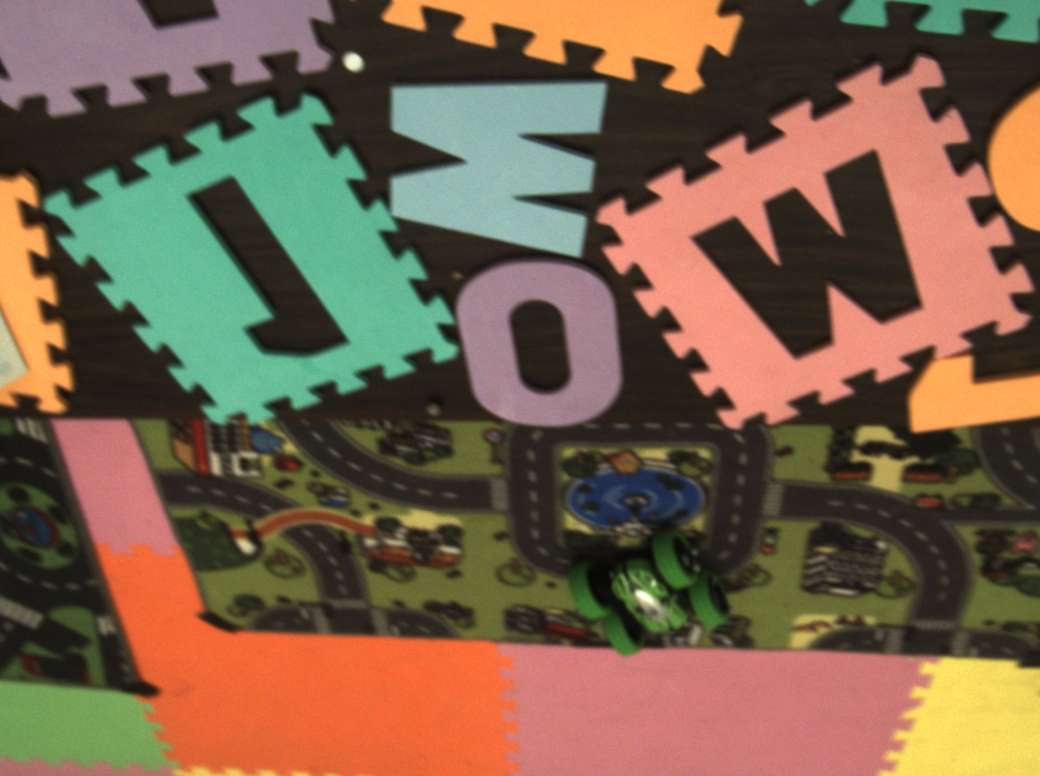} & \includegraphics[width=\linewidth,valign=m]{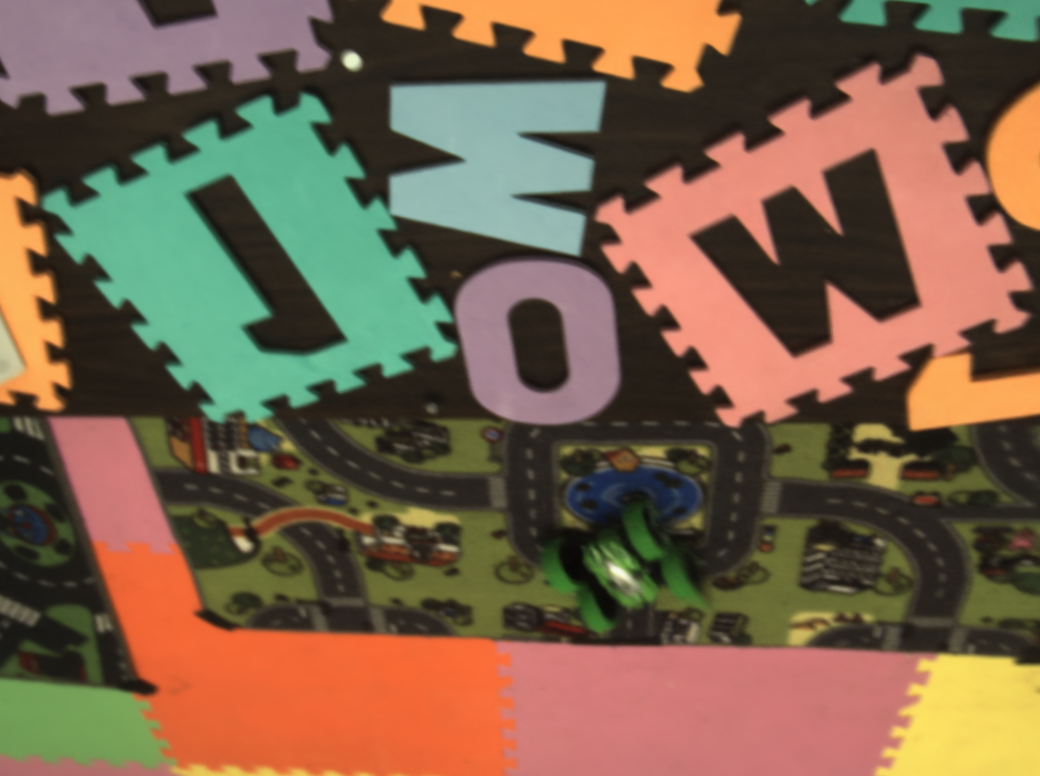} & \includegraphics[width=\linewidth,valign=m]{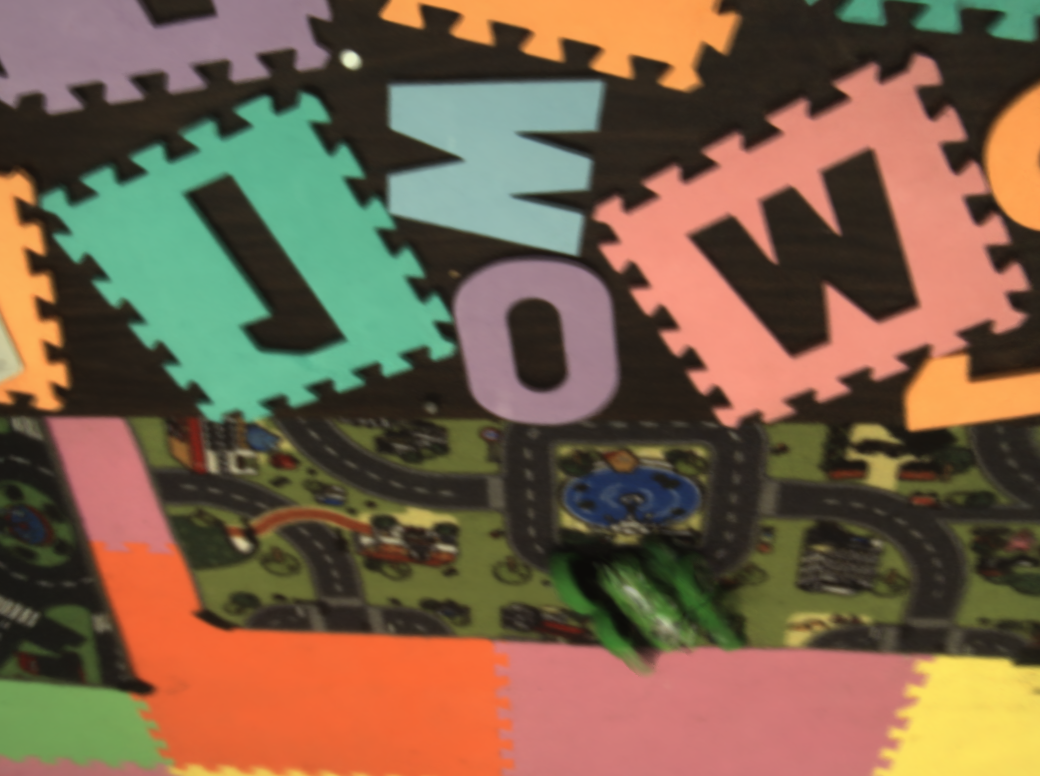} & \includegraphics[width=\linewidth,valign=m]{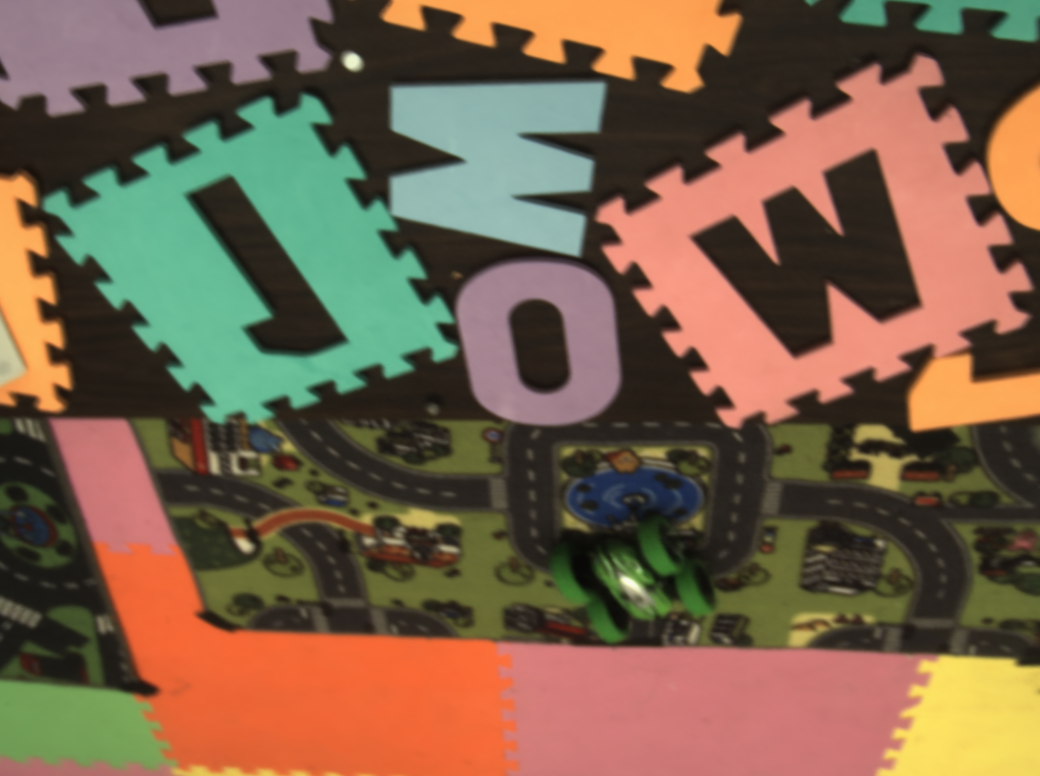} & \includegraphics[width=\linewidth,valign=m]{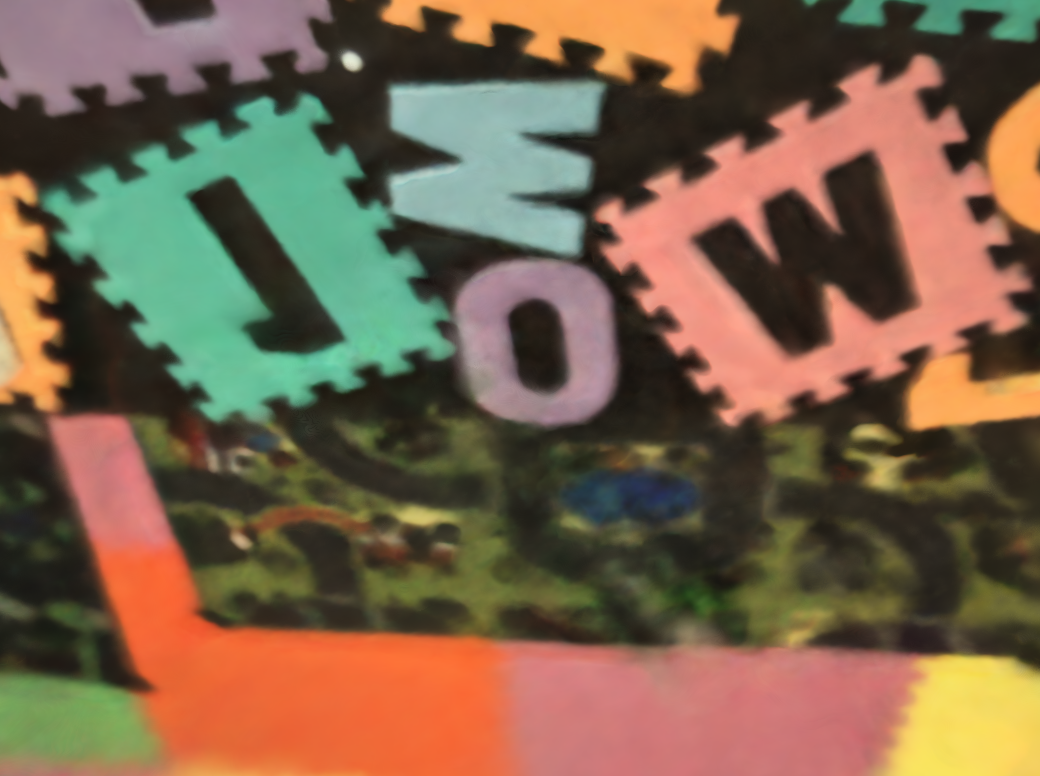} & \includegraphics[width=\linewidth,valign=m]{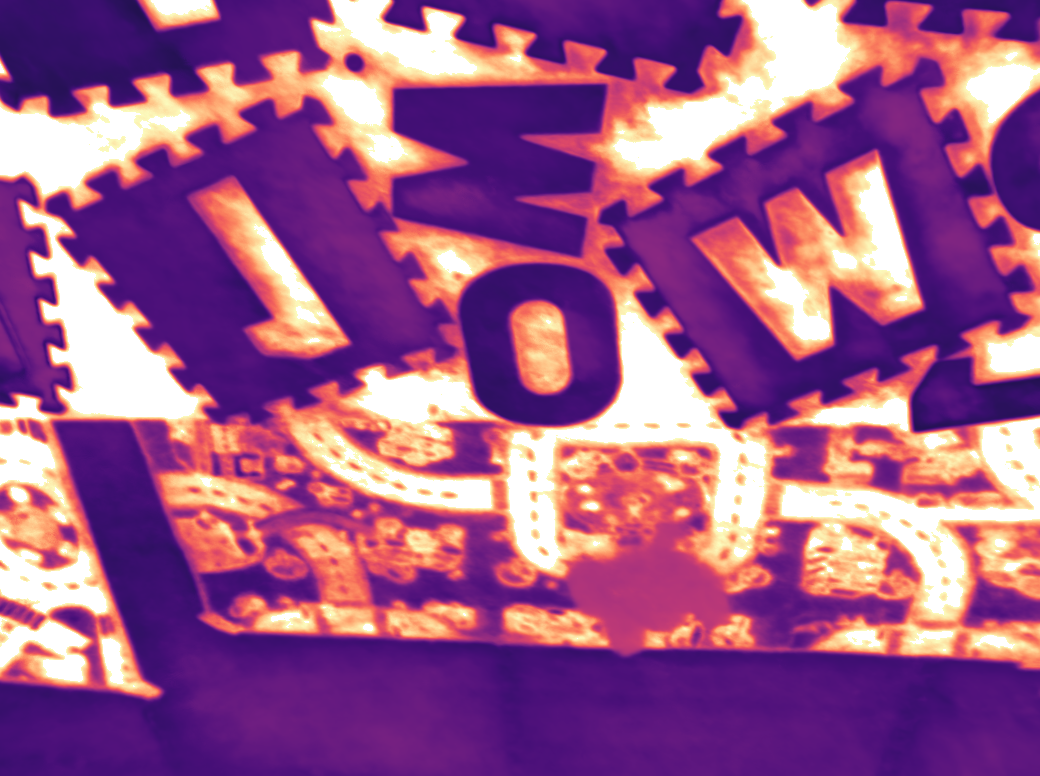 } \\

\rotatebox{90}{UAV} & \includegraphics[width=\linewidth,valign=m]{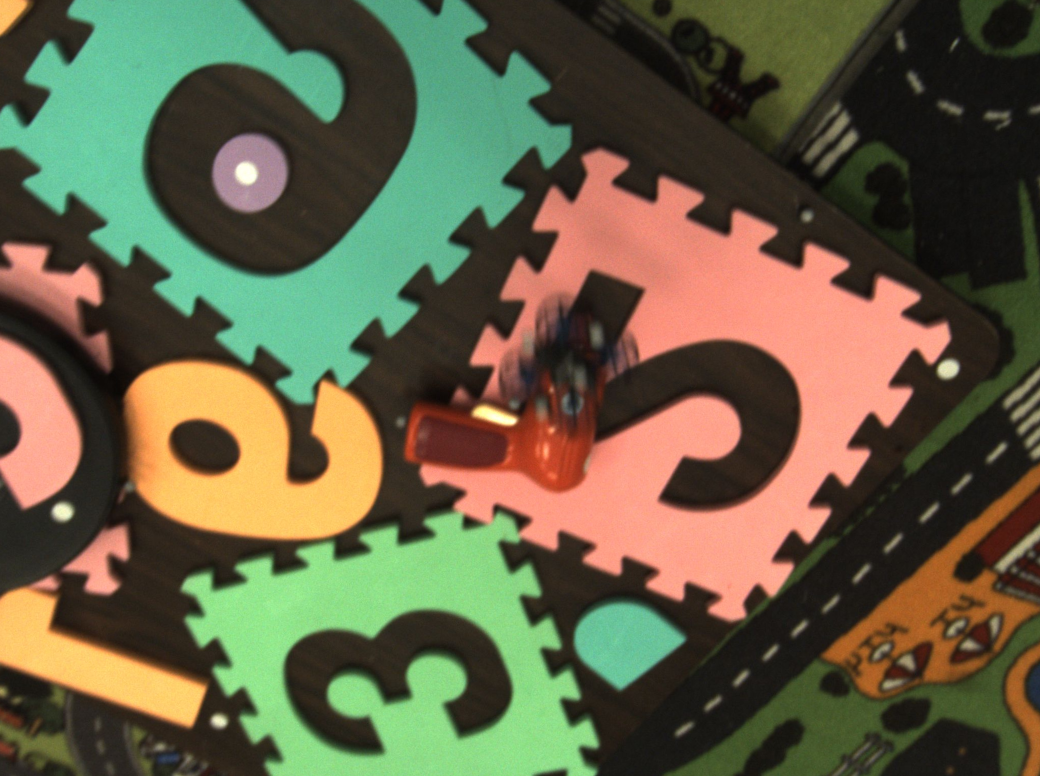} & \includegraphics[width=\linewidth,valign=m]{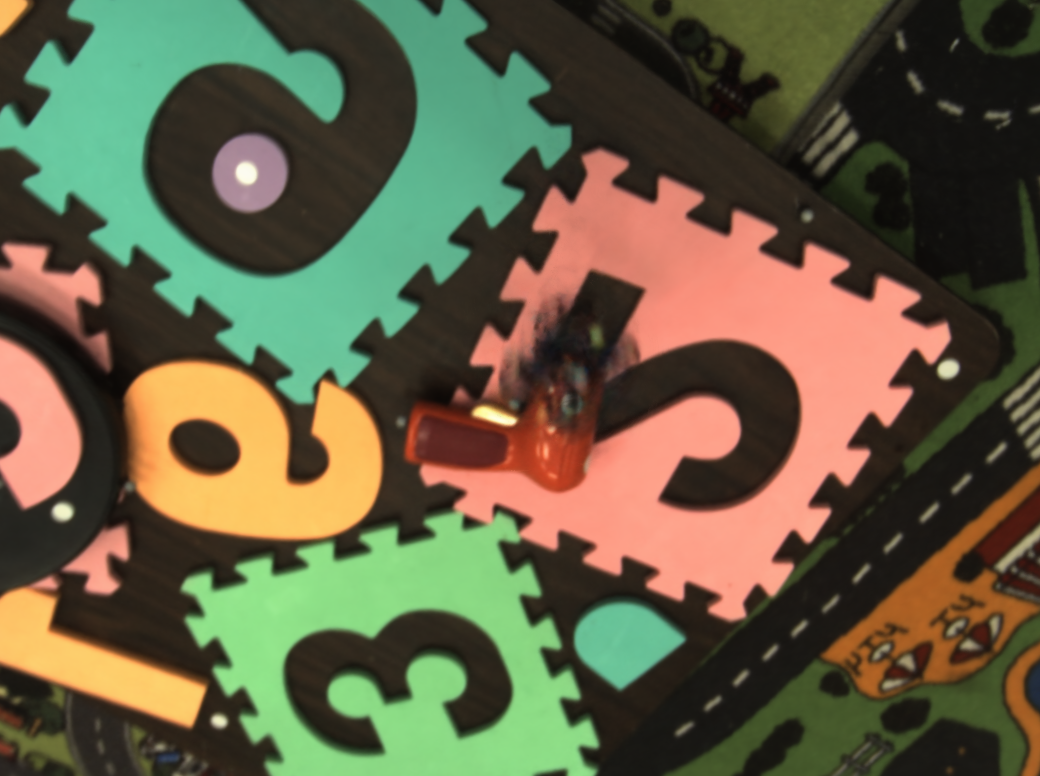} & \includegraphics[width=\linewidth,valign=m]{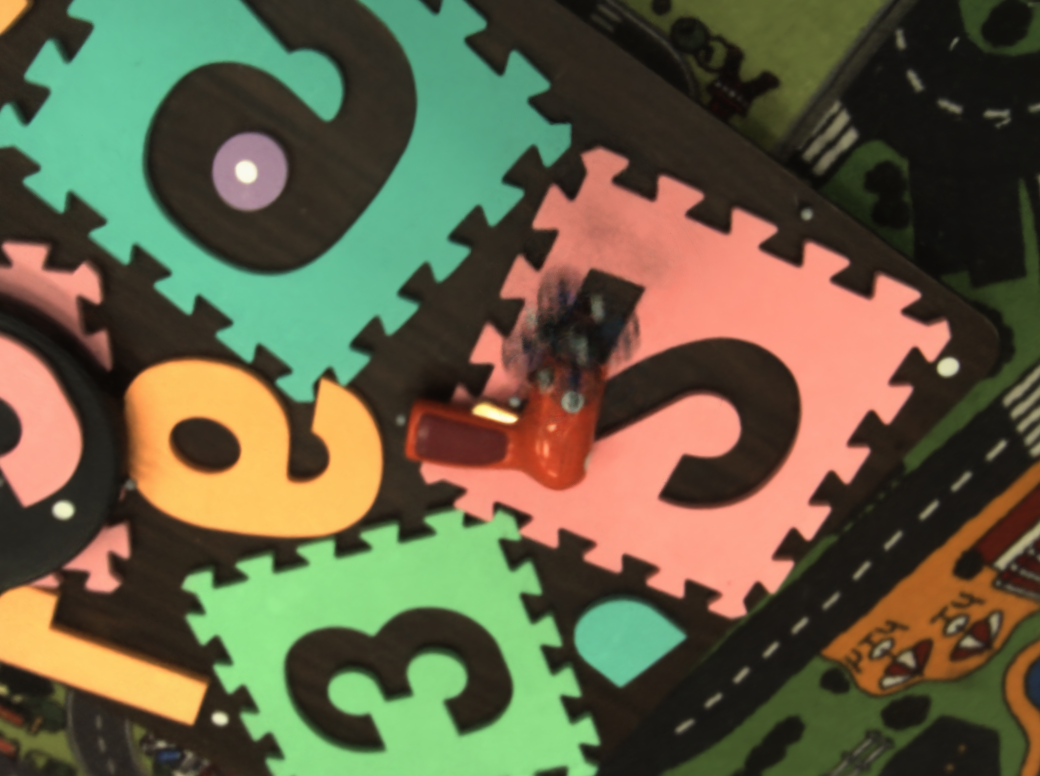} & \includegraphics[width=\linewidth,valign=m]{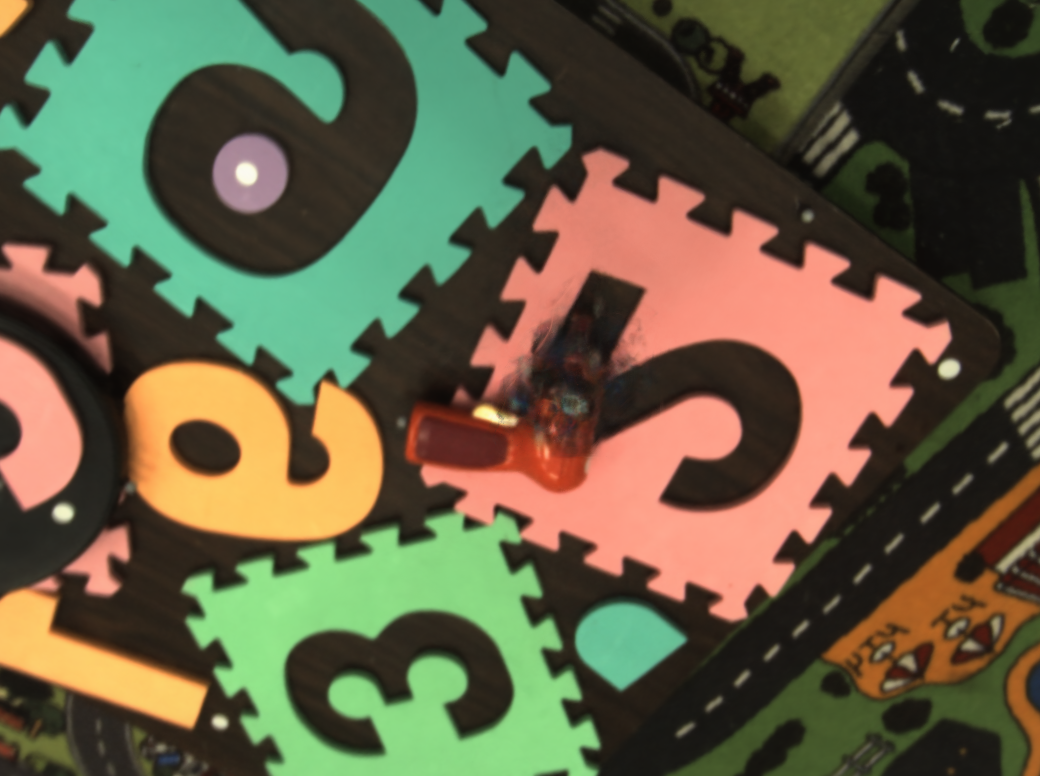} & \includegraphics[width=\linewidth,valign=m]{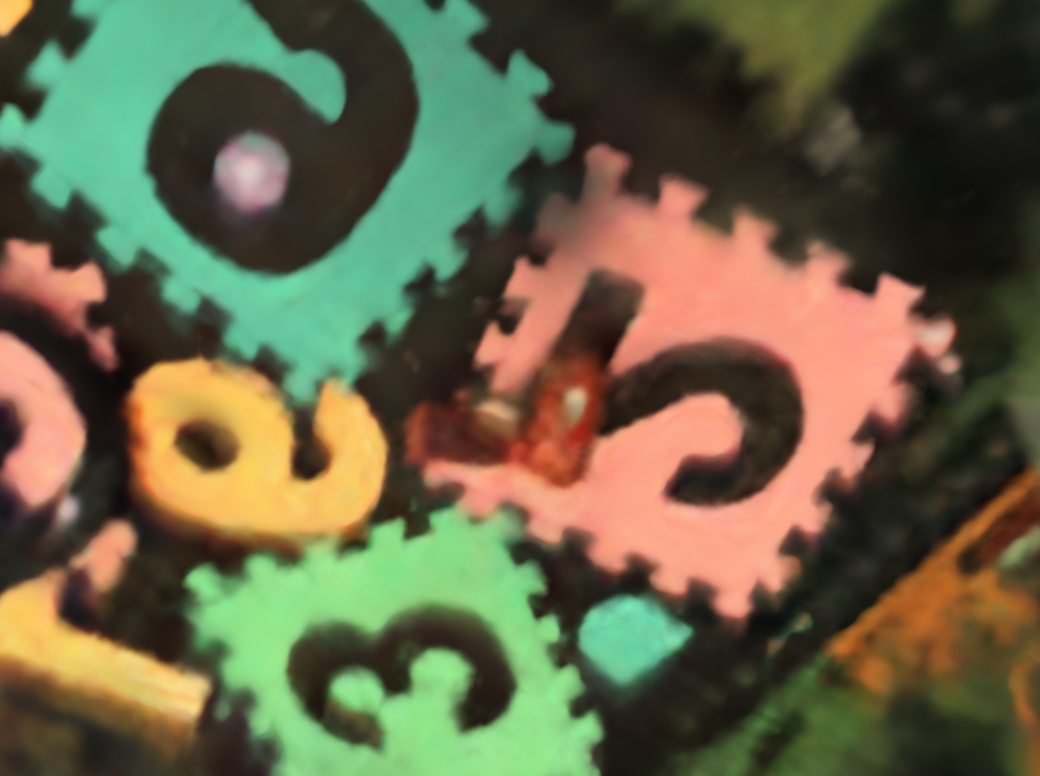} &\includegraphics[width=\linewidth,valign=m]{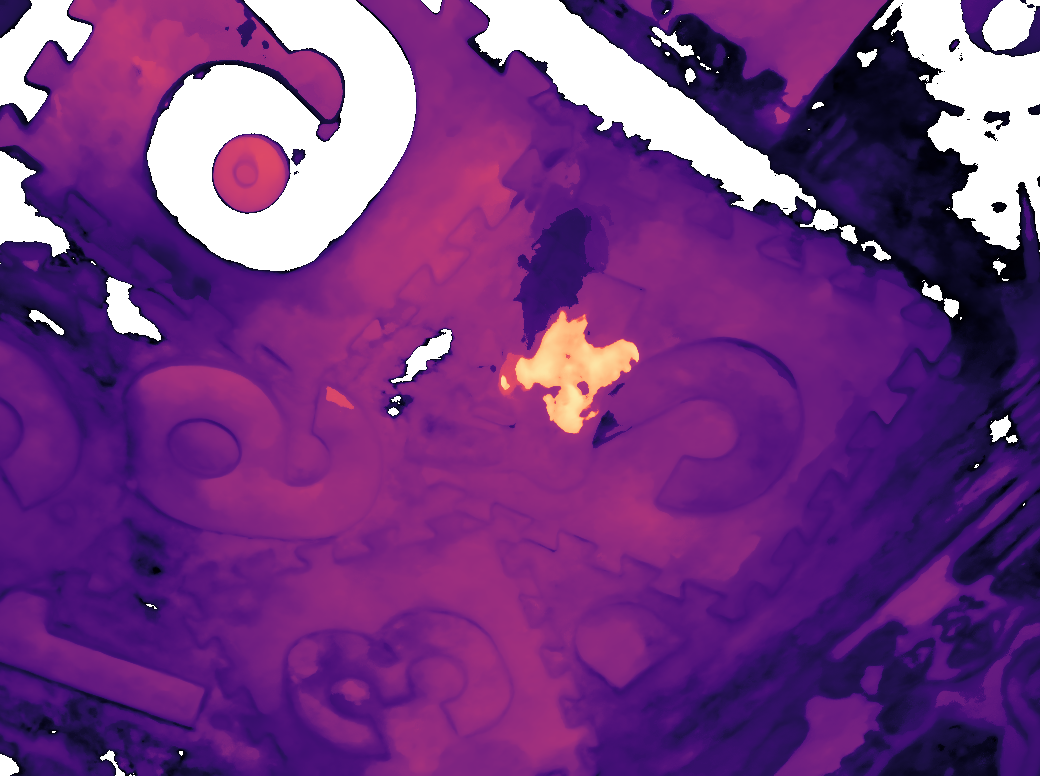}  \\

 & Ground Truth & DE-NeRF(Ours) & DE-Baseline & Nerfies~\cite{park2021nerfies} & NeRF~\cite{mildenhall2021nerf} & Depth (Ours) 
\end{tabular}
\end{center}
\vspace{-0.2cm}
\caption{Qualitative comparisons of our method and baselines on synthetic and real-world datasets.}
\label{fig:qualitative results rgb}
\end{table*}


\begin{table}[t]
\resizebox{\linewidth}{!}{
\begin{tabular}{lcccccc}
\toprule
Number of Views & \multicolumn{2}{c}{10} & \multicolumn{2}{c}{25} & \multicolumn{2}{c}{50}  \\
\cmidrule(r){2-3} \cmidrule(r){4-5} \cmidrule(r){6-7}
Method & PSNR & LPIPS & PSNR & LPIPS & PSNR & LPIPS \\
\midrule
Nerfies\cite{park2021nerfies} & 18.51 & 0.351 & 22.51 & 0.147 & 25.97 & 0.089\\
DE-Baseline & 22.21 & 0.113 & 23.95  & 0.101 & 27.12 & 0.093 \\
\midrule
Ours (no void) & 28.16  & 0.076 & 29.82 & 0.044 & 33.25 & 0.038 \\
Ours & 28.89 & 0.078  & 29.91  & 0.042 & 33.41 & 0.040 \\
Ours + AS & 28.85 & 0.102 & 31.83 & 0.039 & 34.60  & 0.038 \\
Ours + AS + PR & 32.13 & 0.046 &  32.55  & 0.037 & 35.04  & 0.034 \\
\bottomrule
\end{tabular}}
\vspace{1mm}
\caption{\textbf{Ablation Study.} We investigate the effectiveness of void sampling, active sampling (AS) as well as pose refinement (PR). All the proposed components contribute meaningfully to our method. The gain of our method without additional components comes from event integration alone performed using~\eqref{eq:eventLoss}.
}
\label{tab:ablation}
\end{table}

\begin{figure*}
   \centering    
    \includegraphics[width=0.33\linewidth]{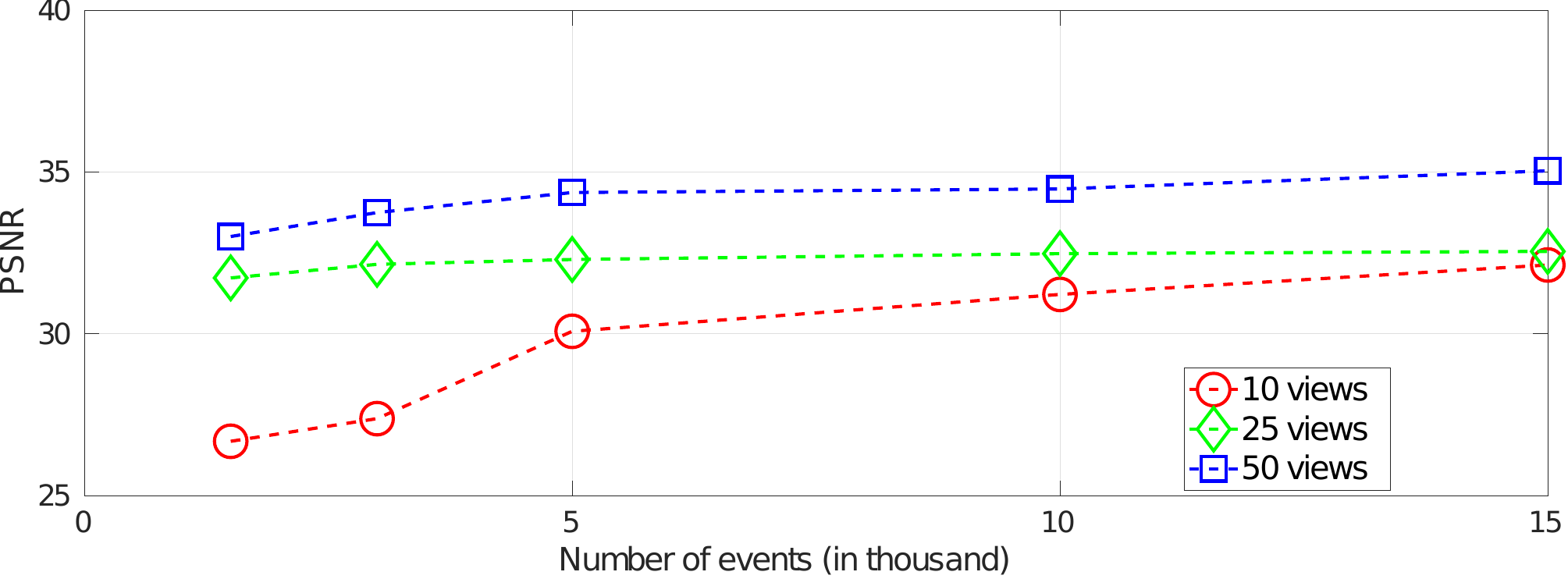}
   \includegraphics[width=0.33\linewidth]{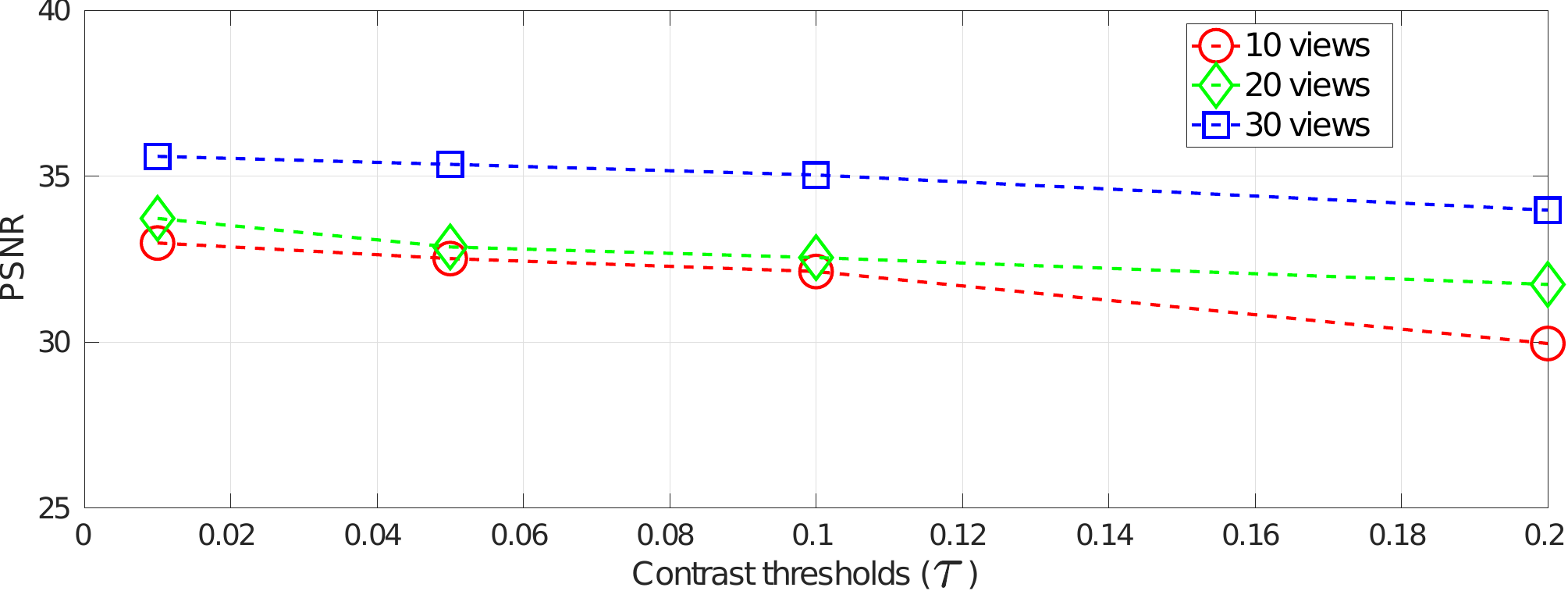}
   \includegraphics[width=0.33\linewidth]{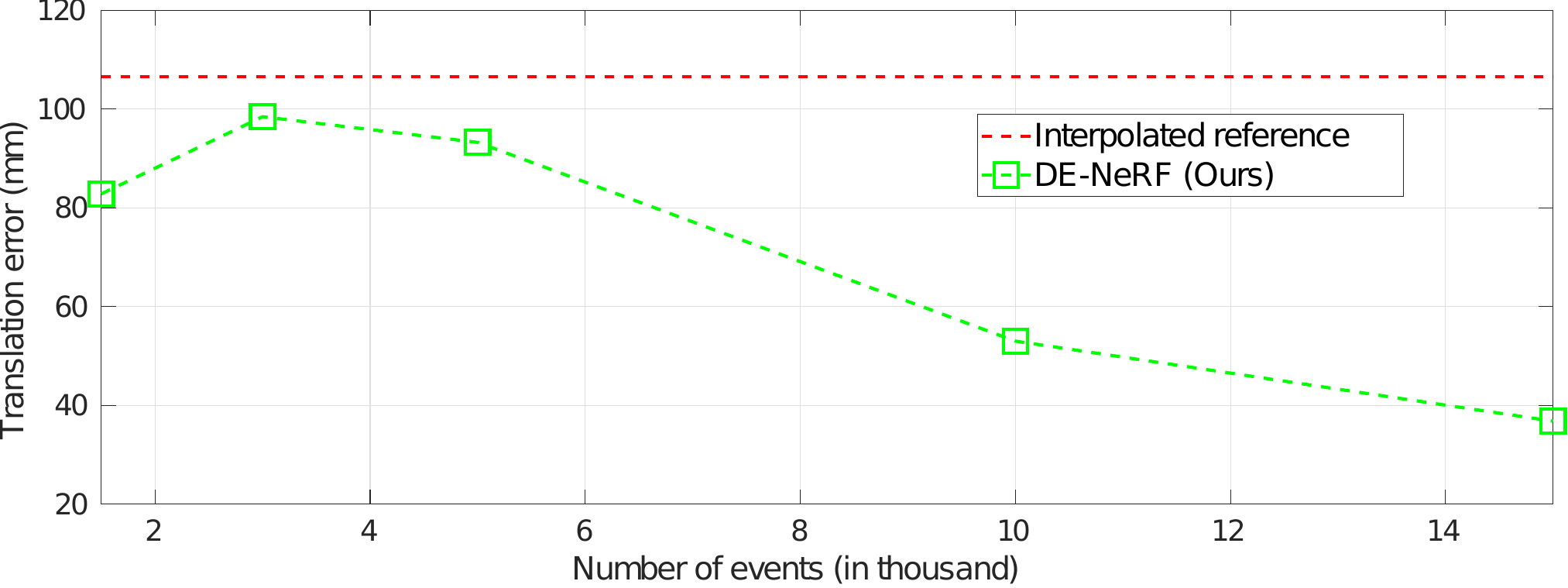}
   \vspace{0.75mm}
   \caption{The behaviour analysis of our method. Novel view synthesis quality vs. RGB views and number of views (left) and even contrast threshold ($\tau$) for sensitivity measure (middle). The pose error measure in absolute translation error vs. number of events used (right). }
   \label{fig:behaviour}
\end{figure*}

\begin{figure}[h!]
\begin{center}
\includegraphics[width=1\linewidth]{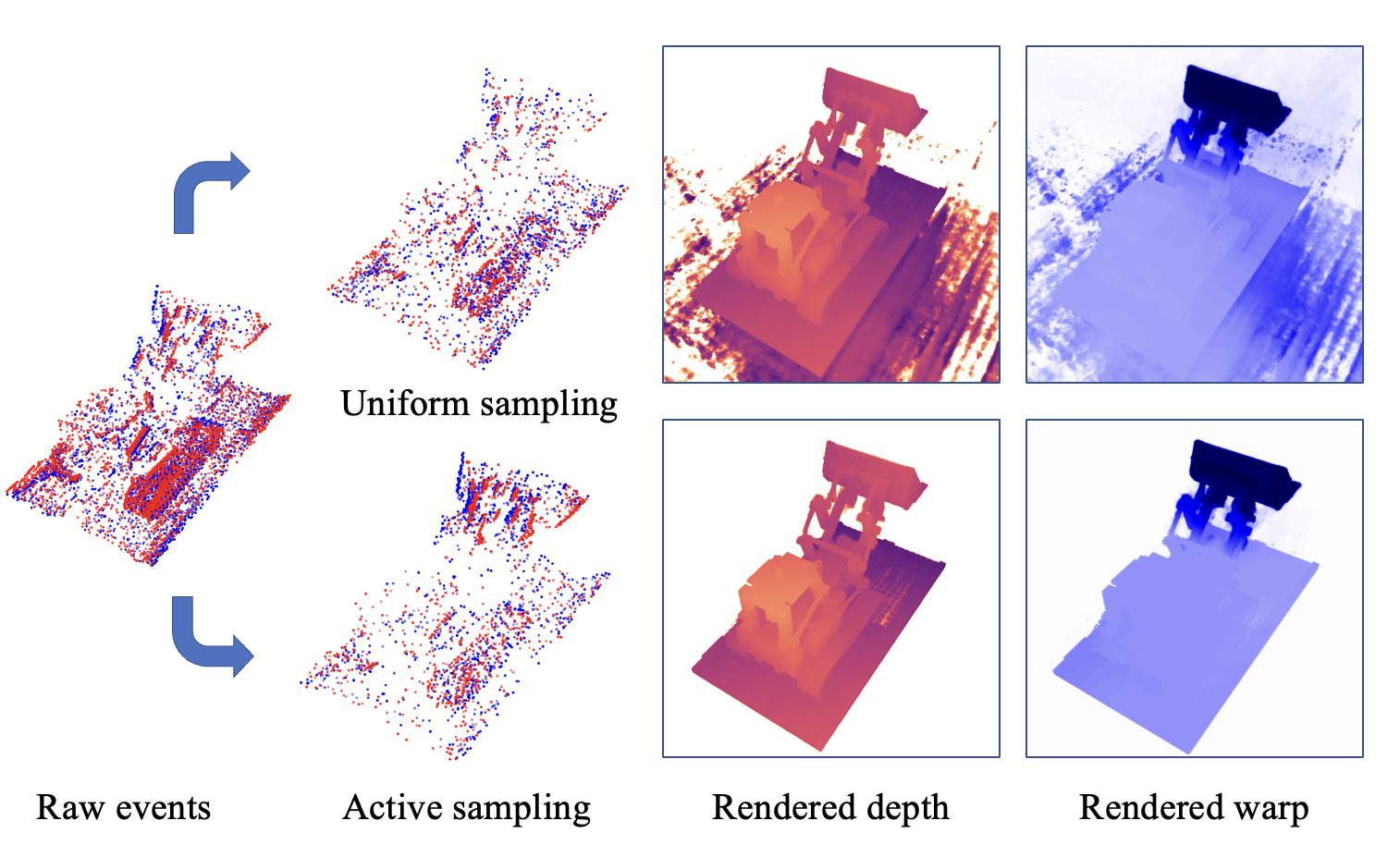}
\end{center}
  \caption{ Uniform vs. active sampling techniques. Our active sampling method uses more events from deformable regions.}
\label{fig:active_sampling}
\end{figure}

In this section, we provide quantitative and qualitative evaluations. In Table~\ref{tab:syt_table} and Table~\ref{tab:real_table}, 
we report the qualitative results for novel view rendering in synthetic and real-world datasets. Our method outperforms all other methods in all MSE, PSNR, SSIM, and LPIPS metrics,  thanks to its ability to model fast-moving deformable scenes. This is particularly highlighted in the Lego, Campfire, and Candle datasets. 
On Campfire, our method not only successfully learns the direction of flame contour changes, with the head of the flame pointing left in the novel view, but also learns the depth changes caused by the flame variations. On Candle, our method correctly learns the flame's depth. Similar improvements in performance can be observed in all datasets. Although Fulid is a very challenging (due to the used deformation prior violation), our method still offers a noticeable improvement. Some qualitative results on various datasets are presented in Figure~\ref{fig:Qualitative_depth} and Table~\ref{fig:qualitative results rgb}. For more visualization please refer to Table 5 in supplementary materials.


The DE-Baseline performs well on synthetic datasets. However, this method needs a large number of samples between the current timestamps and the closest frame. \cite{klenk2023nerf} uses a batch size of 30k pairs of events with NVIDIA A40 which is memory and computationally inefficient. In addition, we found that DE-baseline performs numerically worse than Nerfie across all metrics on most real-world data, but with better visual quality. 
This is similar to the conclusion of the prior work \cite{klenk2023nerf}. 


\begin{table}[h]
\centering
\centering
\resizebox{1.\linewidth}{!}{
\begin{tabular}{c|cc|cc|cc}
\toprule
- & \multicolumn{2}{c|}{E2VID\cite{rebecq2019high}+Nerfies\cite{park2021nerfies}} & \multicolumn{2}{c|}{E2VID+Hyper\cite{park2021hypernerf}} & \multicolumn{2}{c}{E2VID}  \\ 
Dataset  & PSNR  & LPIPS & PSNR & LPIPS & PSNR & LPIPS \\
\hline

Lego & 17.12  & 0.46   & 16.40 & 0.50 &  15.17 & 0.38  \\

Umbrella & 25.05  & 0.44  & 24.93 & 0.46 &  25.92 & 0.15 \\

Selfie  & 18.36 & 0.39 & 17.95 & 0.40 & 16.95 & 0.42\\

\bottomrule
\end{tabular}}
\vspace{0.1cm}
\caption{\textbf{Events-to-frame based method Comparison.} We report results using learning based events-to-frame method E2VID\cite{rebecq2019high}. It can directly synthesize novel view frame using only events. However, as depicted in the third column, the synthesized quality is deficient.}
\label{tab:more-baseline}

\end{table}

Additionally, we provide comparison with events-to-frame method \cite{rebecq2019high}. We reconstruct the intensity image and training them together with RGB images using \ref{eq:rgbLoss}. For static camera setup the background trigger no events so we report results only for the segmented dynamic part. As Table ~\ref{tab:more-baseline} shows the events-to-frame based method exhibits poor performance, primarily attributed to challenges such as unknown absolute intensity during reconstruction, as well as domin shifts and artifacts.

\begin{figure}[t]
\begin{center}
\includegraphics[width=1\linewidth]{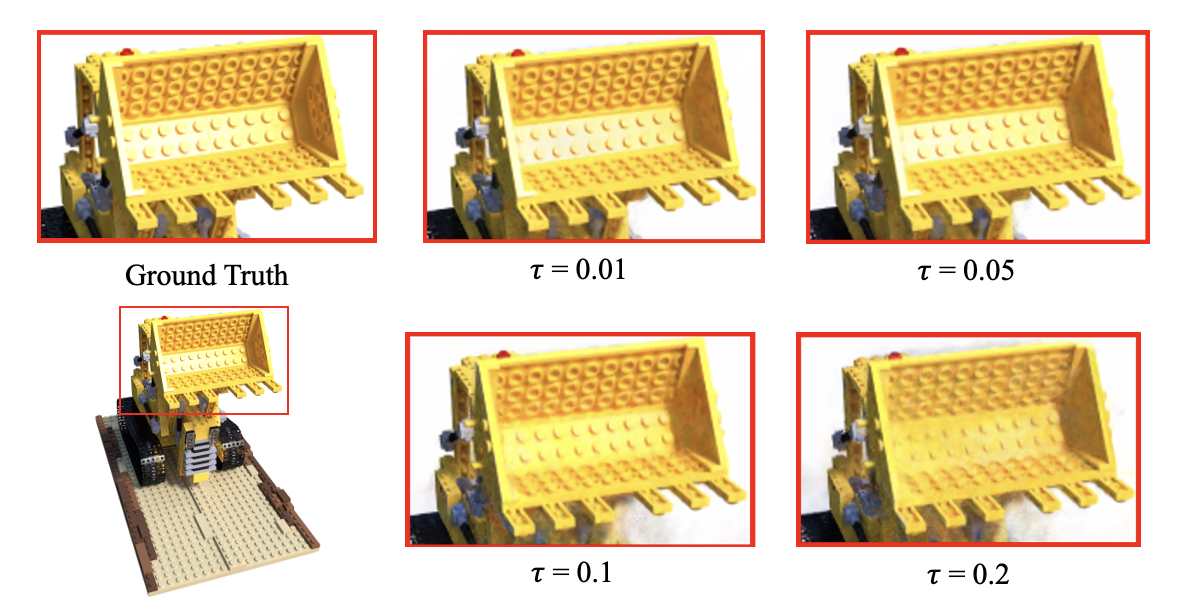}
\end{center}
  \caption{Effect of contrast thresholds on view synthesis. As expected, more sensitive event cameras lead to better representations.}
\label{fig:ct_fig}
\end{figure}

It can be observed that our method has limited improvement on fluid and fountain, which is likely due to the difficulty in simulating fluid solely based on the warping radiance field. Furthermore learning depth for water is also challenging due to the shading effect of water. Our method achieved effectively reconstruction of the left person's head movement on the Selfie dataset.




\subsection{Ablation Study}
We ablate our method for void  sampling, active sampling, and pose refinement. The obtained results are presented in Table~\ref{tab:ablation}. Our findings indicate that solely relying on with-event location sampling leads to a slight decline in performance. This result may be attributed to the small sampling window utilized in our study (the entire trajectory was divided into 200 windows)
, which necessitates that with-event methods provide sufficient information for brightness changes over time. Additionally, our results show that active sampling improves the performance in experiments with 25 and 50 RGB views, with a minimal effect for  10 views. This is expected because for sparse views, the main source of error is triggered by bad events pose estimation from interpolation. As Figure~\ref{fig:active_sampling} shows adopting active sampling allows for taking advantage of more events triggered by deformation, thereby efficiently learning the warp field. Compared to random sampling, our method achieves more accurate depth in the 10 views case. Finally, our results demonstrate that the use of pose refinement techniques enhances the performance for 10 views cases and leads to further improvements for 25 and 50 views.

\subsection{Behaviour Analysis}
\begin{figure}[t]
    \centering
    \includegraphics[width=0.960\linewidth]{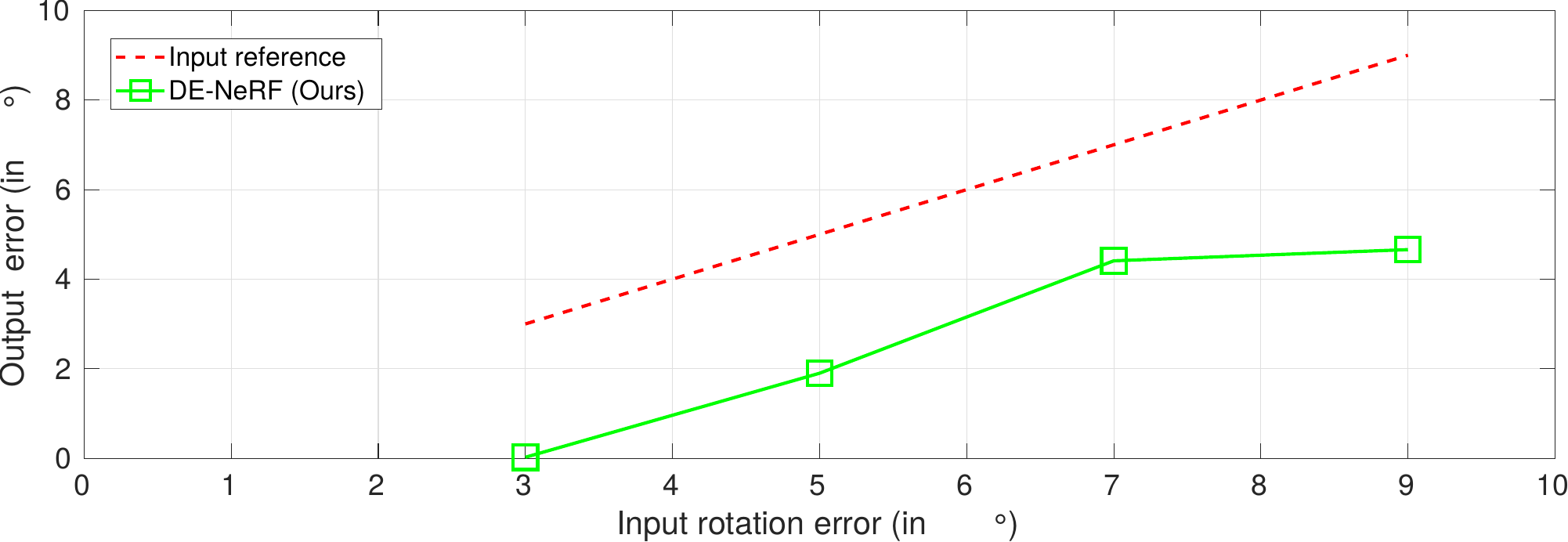}    
    \caption{{PoseNet robustness against injected rotation noise.} }
    \label{fig:roation_error}
\end{figure}

We conduct several experiments on Lego to investigate the behavior of our method. 
The performed studies are summarized in Figure~\ref{fig:behaviour}, with some graphical illustrations in Figure~\ref{fig:ct_fig}. 
It can be observed that the increasing number of events impacts positively the novel view synthesis as well as the pose estimation, in all cases. At the same time, the lower contrast threshold, or higher sensitivity of the event camera, also leads to better performance, as expected. The pose error in Figure~\ref{fig:behaviour} (right) is evaluated using the ATE-RMSE\cite{zhu2022nice}. We also provide the error obtained by initial pose interpolation, for the reference. In Figure ~\ref{fig:roation_error} we report the rotation error of our method after injecting different magnitudes of rotation noise. We found that our method is robust to small rotation noise and can effectively reduce large rotation noise.

\section{Conclusion}
In this work, we demonstrated the benefits of event cameras in modelling fast deforming radiance fields. The success of our method is contributed by the proposed novel neural architecture design, training strategy, and the instantaneous nature of the asynchronous event streams. Our extensive experiments on diverse real and synthetic datasets revealed very exciting results with significant performance gain, both quantitatively and qualitatively. The success of the proposed method must also be credited to the recent advancements in radiance field modeling. This is particularly the case because the integration of event cameras  in the radiance field is in fact very natural. This allowed us to quickly establish a baseline and improve it using the techniques proposed in this paper. Our method opens new avenues for the 3D visual modeling of fast-moving cameras and deforming scenes, in a relatively simple manner. 

\paragraph{Limitations:} For monochromatic events, our method occasionally generates color artifacts. Our method benefits insignificantly in very complex scenes that largely violate the assumed deformation model. This can be seen with the Fluids dataset. We believe this limitation can be addressed by more sophisticated non-rigid priors for complex scenes.

\paragraph{Acknowledgements:} Research is partially funded by VIVO Collaboration Project on Real-time scene reconstruction and also partially funded by the Ministry of Education and Science of Bulgaria (support for INSAIT, part of the Bulgarian National Roadmap for Research Infrastructure).

{\small

\bibliographystyle{ieee_fullname}
\bibliography{egbib}
}

\end{document}